\documentclass[article,10pt]{IEEEtran}

\usepackage[utf8]{inputenc} 
\usepackage[T1]{fontenc}
\usepackage{url}              
\usepackage{cite}             

\usepackage[cmex10]{amsmath} 
\usepackage{mleftright}      
\mleftright                  
\usepackage{amsmath}
\DeclareMathOperator*{\argmin}{argmin}
\usepackage{graphicx}         
\usepackage{booktabs}         
\usepackage{mathtools}
\newcommand\myeq{\stackrel{\mathclap{\small\mbox{def}}}{=}}     
\usepackage{graphicx,cite,amsmath,amssymb,amsfonts,mathtools,algorithm,mathabx,textcomp,blindtext,xcolor,subfig,float,mwe,dblfloatfix,bm, xfrac,algpseudocode, stackengine, verbatim,xparse} 
\usepackage{amsthm}

\newtheorem{proposition}{Proposition}
\newtheorem{corollary}{Corollary}

\begin{document}
\title{Multi-Timescale Ensemble $Q$-learning for Markov Decision Process Policy Optimization 
	\thanks{Talha Bozkus and Urbashi Mitra are with the Ming Hsieh Department of Electrical and Computer Engineering, University of Southern California, Los Angeles, USA. Email: \{bozkus, ubli\}@usc.edu. }
	\thanks{This work was funded by the following grants: ARO W911NF1910269, DOE DE-SC0021417, Swedish Research Council 2018-04359, NSF CCF-2008927, NSF RINGS-2148313, NSF CCF-2200221, NSF CIF-2311653, ONR 503400-78050, ONR N00014-15-1-2550 and USC + Amazon Center on Secure and Trusted Machine Learning}
}

\author{Talha Bozkus and Urbashi Mitra}
	
\maketitle

\begin{abstract}
Reinforcement learning (RL) is a classical tool to solve
network control or policy optimization problems in unknown
environments. The original $Q$-learning suffers from performance and complexity challenges across very large networks. Herein, a novel model-free ensemble reinforcement learning algorithm which adapts the classical $Q$-learning is proposed to handle these challenges for networks which admit Markov decision process (MDP) models. Multiple $Q$-learning algorithms are run on multiple, distinct, synthetically created and structurally related Markovian environments in parallel; the outputs are fused using an adaptive weighting mechanism based on the Jensen-Shannon divergence (JSD) to obtain an approximately optimal policy with low complexity. The theoretical justification of the algorithm, including the convergence of key statistics and $Q$-functions are provided. Numerical results across several network models show that the proposed algorithm can achieve up to 55\% less average policy error with up to 50\% less runtime complexity than the state-of-the-art $Q$-learning algorithms.  Numerical results validate assumptions made in the theoretical analysis.
\end{abstract}

\section{Introduction}\label{sec:introduction}
Markov Decision Processes (MDPs) are natural mathematical tools for modeling sequential decision-making problems in many large real-world networks \cite{mdp1,mdp2,mdp3}. When the underlying system dynamics are observable, the optimization problem of MDPs can be solved by dynamic programming \cite{bertsekas_book}. However, these algorithms are not directly applicable to problems where the underlying MDP is unknown (or non-observable), in which case \textit{model-free} Reinforcement Learning (RL) algorithms such as $Q$-learning can be employed to simulate the system dynamics and learn the policies and value functions \cite{barto_sutton_rl}. 

$Q$-learning can be employed to solve a variety of optimization and control problems in unknown environments \cite{q_learning_tsp_1, q_learning_tsp_2, q_learning_tsp_3, q_learning_tsp_4}. However, it suffers from several performance and complexity challenges across large MDPs, including high estimation bias and estimation variance, training instability, slow convergence, and high sample complexity. To this end, several variants of $Q$-learning have been developed to handle these challenges. Estimation bias is considered in \cite{double_q, weighted_q_learning}, and the estimation variance and training stability are examined in \cite{averaged_dqn, randomized_double_q}. The convergence rate is improved in \cite{speedy_q}, and training data efficiency is considered in \cite{fitted_q}. In \cite{liu2020sampled,libin_paper}, strategies specific to wireless networks are considered.  As these algorithms have different objectives, their strategies and implementations also differ. Similar to the original $Q$-learning, a single $Q$-function estimator is employed in \cite{speedy_q, fitted_q}. On the other hand, multiple $Q$-function estimators are used in \cite{double_q, maxmin-q, averaged_dqn, ensemble_bootstrap_q, randomized_double_q}, in which each estimator is initialized independently, and their outputs are fused into a single estimate via a weighting mechanism. These algorithms directly operate on the original Markovian environment; however, $Q$-learning algorithms using multiple $Q$-function estimators on {\bf multiple} Markovian environments have not been well-studied. We will see that the performance and complexity of $Q$-learning can be further improved by employing multiple $Q$-function estimators on multiple structurally related synthetic Markovian environments operating at different time-scales. 

Despite extensive prior work \cite{survey1, survey2}, achieving efficient and scalable exploration in large Markovian environments remains a major challenge in reinforcement learning. \textit{Too little exploration} may cause the agent to behave greedily. Consequently, some parts of the environment may never be visited, and the agent may keep selecting sub-optimal actions and get stuck in a local optimum. On the other hand, \textit{too much exploration} may yield a high accumulated cost, preventing the utilization of previous experiences, and be computationally expensive for very large environments. While the previous work generally employs a \textit{single} efficient exploration strategy \cite{survey1, survey2}, we herein propose a \textbf{two-level exploration} strategy, \textit{i.e.} there are two different sources of exploration: at the \textit{algorithm-level}, we use one of the existing exploration techniques (such as epsilon-greedy $Q$-learning \cite{barto_sutton_rl}), and at the \textit{environment-level}, we utilize multiple, distinct, synthetically created and structurally related Markovian environments, which provide different orders of relationships between states. As we consider products of the probability transition matrix of the original system to construct our synthetic systems, we deem our approach as having {\bf multiple time scales}, corresponding to the $n$-hop transition matrices of the Markov chain. Our goal is to improve the exploration capabilities of the agent and accelerate the exploration stage of $Q$-learning.

To this end, we propose a novel ensemble $Q$-learning algorithm, where \emph{multiple} $Q$-learning algorithms are run in parallel on \emph{multiple} distinct, synthetically created and structurally related Markovian environments. Their outputs are fused into a single $Q$-function estimate using an adaptive weighting mechanism based on a Jensen–Shannon divergence between the distributions corresponding to the $Q$-functions of different environments. In the end, an approximately optimal deterministic policy is obtained with low complexity. In our prior work \cite{talha_eusipco, talha_icassp}, we introduced similar algorithms and presented preliminary findings. That initial analysis yielded significant insights into the advantages of employing multiple Markovian environments to improve the accuracy and complexity of the original $Q$-learning.  Herein, we provide significant improvements over that preliminary work: (i) We propose a more interpretable and computationally cheaper way to construct multiple synthetic Markovian environments. (ii) We remove the constraints and approximations on the structure of the network models; hence, our current design is applicable to a wider range of networks. (iii) We provide a more complete theoretical analysis of the proposed algorithm. (iv) We utilize a new distance metric based on the $\operatorname{JSD}$ to improve the accuracy of the adaptive weighting mechanism.

The main contributions of the paper are as follows: (i) We systematically construct the multiple synthetic Markovian environments to enable an efficient and scalable exploration in $Q$-learning. (ii) We propose a novel $Q$-learning algorithm based on an ensemble of the $Q$-functions from multiple Markovian environments. (iii) We provide theoretical analyses on the convergence and error variance of the proposed algorithm. (iv) We simulate the algorithm on several large real-world network classes. Numerical results show that the proposed method outperforms the state-of-the-art $Q$-learning algorithms on all networks, achieving up to 55\% less average policy error with up to 50\% less runtime complexity. In addition, simulations confirm the theoretical analyses.

We use the following notation: the vectors are bold lower case (\textbf{x}), matrices and tensors are bold upper case (\textbf{A}), and sets are in calligraphic font ({$\mathcal{S}$}).

\section{System Model and Tools}
\label{sec:system_model}

\subsection{Infinite Horizon Discounted Cost MDP model}
\label{subsec:mdp}
MDPs are characterized by 4-tuples $\{\mathcal{S}$, $\mathcal{A}$, $p$, $c$\}, where $\mathcal{S}$ and $\mathcal{A}$ denote the finite state and action spaces, respectively. We denote $s_{t}$ as the {\em state} and $a_{t}$ as the {\em action} taken at discrete time period $t$. The transition from state $s$ to $s^{\prime}$ under action $a$ occurs with probability $p_{a}(s,s^{\prime})=p(s^{\prime} = s_{t+1} \mid s = s_{t}, a = a_{t})$, which is stored in the $(s,s',a)^{th}$ element of the three-dimensional probability transition tensor (PTT) $\mathbf{P}$, and a bounded average cost $c_{a}(s)= \sum_{s^{\prime} \in \mathcal{S}} p_{a}\left(s, s^{\prime}\right) \hat{c}_{a}(s, s^{\prime})$ is incurred, which is stored in the $(s,a)^{th}$ element of the cost matrix $\mathbf{C}$, where $\hat{c}_{a}(s, s^{\prime})$ is the instantaneous transition cost from state $s$ to $s^{\prime}$ under action $a$. We denote the probability transition matrix (PTM) and cost vector under the action $a$ by $\mathbf{P}_a$ and $\mathbf{c}_a$, respectively. We focus on infinite horizon discounted cost MDPs, where $t = \mathbb{Z}^{+}\cup \{0\}$. Our goal is to solve \textit{Bellman's optimality} equation:  
\begin{align}
\mathbf{v}^{*}(s)&=\min _{\pi} \mathbf{v}_{\pi}(s)=\min _{\pi}\mathbb{E}_{\pi}\left[\sum_{t=0}^{\infty} \gamma^{t} c_{a_{t}}(s_{t}) | s_{0}=s\right],\label{Equ: optimization_eq}\\
\mathbf{\pi}^*(s)&=\argmin_{\pi} \mathbf{v}_{\pi}(s),\label{Equ: optimization_eq_2}
\end{align}
for all $s \in \mathcal{S}$, where $\mathbf{v}_{\pi}$ is the \textit{value function} \cite{bertsekas_book} under the \textit{policy} ${\pi}$, $\mathbf{v}^{*}$ is the \textit{optimal value function}, $\mathbf{\pi}^*$ is the \textit{optimal policy}, and $\gamma \in (0,1) $ is the discount factor. The policy $\pi$ can define either a specific action per state (\textit{deterministic}) or a distribution over the action space per state (\textit{stochastic}) for each time period. If the policy does not change over time, \textit{i.e.,}  $\pi_{t} = \pi,$ $\forall t$, then it is deemed \textit{stationary}. There always exists a deterministic stationary policy that is optimal given a finite state and action spaces \cite{bertsekas_book}. Hence, we, herein, consider deterministic and stationary policies.

\subsection{$Q$-Learning}\label{subsec:q_learning}

When the system dynamics ($p$ and $c$) are unknown or non-observable, $Q$-learning can be used to solve (\ref{Equ: optimization_eq}) and (\ref{Equ: optimization_eq_2}). $Q$-learning seeks to find the optimal policy $\pi^{*}$ by learning the $Q$ functions for all $(s,a)$ pairs using the following update rule:
\begin{equation}\label{Equ: Q-learning-update-rule}
    Q(s, a) \leftarrow(1-\alpha) Q(s, a)+\alpha(c_{a}(s)+\gamma \min _{a' \in \mathcal{A}} Q(s', a')),
\end{equation}
where $\alpha \in (0,1)$ is the learning rate. In practice, $\epsilon$\textit{-greedy} policies are used to tackle the \textit{exploration-exploitation} trade-off to ensure that sufficient sampling of the system is captured by visiting each state-action pair sufficiently many times \cite{barto_sutton_rl}. To this end, a random action is taken with probability $\epsilon$ (exploration), and a greedy action that minimizes the $Q$-function of the next state is taken with probability $1-\epsilon$ (exploitation). The agent interacts with the environment and collects samples $\{s,a,s',c\}$ to update $Q$-functions using (\ref{Equ: Q-learning-update-rule}). The learning strategy must specify the trajectory length ($l$) (the number of states in a trajectory) and the minimum number of visits to each state-action pair ($v$), which is generally used as a termination condition for the sampling operation. $Q$-functions converge to their optimal values with probability one, \textit{i.e.,} $Q(s,a) \xrightarrow{\mathit{w.p.1}} Q^{*}(s,a)$ for all $(s,a)$ if necessary conditions are satisfied \cite{q-learning-convergence}. The optimal policy and value functions can be inferred from the $Q$-functions as follows:
\begin{align}
\pi^*(s)=\argmin_{a \in \mathcal{A}} Q^{*}(s, a), \hspace{10pt} \mathbf{v}^*(s)=\min_{a \in \mathcal{A}} Q^{*}(s, a).
\end{align}

\subsection{Prior work on Ensemble $Q$-learning \& Model Ensembles}

There has been extensive work on \textit{ensemble $Q$-learning} algorithms in the RL literature. For instance, \cite{ensemble_bootstrap_q} extends double $Q$-learning \cite{double_q} to reduce estimation bias through multiple estimators, while \cite{bootstrapped_error_Q} enhances training stability by reducing error accumulation based on bootstrapping error. In \cite{agarwal2020optimistic}, a robust $Q$-learning method employs random convex combinations of multiple $Q$-functions, and \cite{averaged_dqn} reduces approximation error variance through $Q$-function averaging for more stable training. A randomized ensemble double $Q$-learning is presented in \cite{chen2021randomized} to improve sample efficiency, while \cite{lee2022offline} employs pessimistically trained offline $Q$-functions for the same purpose.

A variety of works have also leveraged \textit{model ensembles} for a variety of RL problems. For example, \cite{modi2020sample} approximates the real environment using a linear combination of pre-trained models. To improve sample complexity, \cite{kurutach2018model} utilizes an ensemble of deep neural networks, \cite{chua2018deep} employs an ensemble of bootstrapped models encoding probability distributions, and \cite{yao2021sample} introduces an ensemble of Bayesian neural network-based dynamics models. Moreover, \cite{pathak2019self} improves exploration via model disagreement based on uncertainty estimates from ensembles.

Despite this prior work, $Q$-learning algorithms using \textit{multiple} $Q$-function estimators on \textit{multiple} Markovian environments (i.e., multiple models) have not been well-studied. By combining the power of ensemble $Q$-learning and model ensembles, we will see that sampling and training can be accelerated, and more accurate and stable $Q$-functions can be produced.

\subsection{Sampling \& Creating Multiple Markovian Environments}\label{subsec:multiple_markovian}

\algdef{SE}[REPEATN]{RepeatN}{End}[1]{\algorithmicrepeat\ #1 \textbf{times}}{\algorithmicend}
\begin{algorithm}[t]
\caption{Sampling \& Creating Multiple Environments}
\hspace*{\algorithmicindent} \textbf{Input:} $l$, $v$, $K$, $\mathcal{M}^{(1)}$ \\
\hspace*{\algorithmicindent} \textbf{Output:} $\mathcal{M}^{(n)}$ for $n \in \{2, 3, \ldots, K\}$
\begin{algorithmic}[1]
    \State Initialize each element of $\mathbf{\hat{P}}_a$ with $\frac{1}{|\mathcal{S}|}$ for $a \in \mathcal{A}$
    \While{each $(s,s')$ in $\mathcal{M}^{(1)}$ not experienced $v$ times}
        \State Choose an initial state $s$ randomly from $\mathcal{S}$ 
        \RepeatN{$l$}
            \State Sample $\{s, a, s'\}$ from $\mathcal{M}^{(1)}$
            \State $\mathbf{\hat{P}}_a(s,s') \gets \mathbf{\hat{P}}_a(s,s') + 1$
        \End
    \EndWhile
    \State Normalize the sum of each row in $\mathbf{\hat{P}}_a$ to 1 for $a \in \mathcal{A}$
    \For{$n \in \{2, 3, \ldots, K\}$}
        \State Create $\mathbf{\hat{P}}^n_a$ by taking $n^{th}$ matrix power of $\mathbf{\hat{P}}_a$ for $a \in \mathcal{A}$
        \State Denote $\mathcal{M}^{(n)}$ as the synthetic Markovian environment corresponding to the $\mathbf{\hat{P}}^n$ 
    \EndFor 
\end{algorithmic}
\label{Algorithm:sampling_creating_multiple_ptts}
\end{algorithm}

There are several ways to create multiple environments (and the corresponding PTTs) based on the PTT of the original environment $\mathbf{P}$. A natural strategy is to employ some function of $\mathbf{P}$ (or $\mathbf{P}^T$). In particular, the probability $p_a(s,s')$ should be related across different environments, and the PTTs of different environments should be row-stochastic {($\textit{i.e.}$ each PTM corresponding to a different action is row-stochastic)} or can be converted into row-stochastic environments by employing appropriate normalization without changing the original structure. We herein propose to use $n$-hop PTTs ($\mathbf{P}^n$) because (i) they describe the $n$-step transition probabilities between states; hence, they are interpretable, corresponding to multiple time-scales, (ii) they are easily computable using efficient matrix multiplication methods, (iii) they are inherently row-stochastic, and (iv) they lead to a nice mathematical analysis (as will be shown later).

There are various factors that suggest employing n-hop Markovian environments could improve the exploration capabilities of the overall system in several ways: (i) They enable the agent to traverse longer trajectories and uncover new state-action pairs beyond its immediate reach, potentially expediting the agent's understanding of the environment with fewer interactions. (ii) They enable the agent to learn from indirect experiences by simulating trajectories that are not directly observed. (iii) They can encourage the agent to consider longer-term rewards and take actions that may not have immediate rewards, leading to better long-term performance, particularly in environments with sparse rewards or long-term dependencies. (iv) They enable the agent to exploit environment patterns by exploring longer trajectories that uncover complex relationships between actions and outcomes, which can be particularly valuable in structured or repetitive environments such as mazes or puzzles, where the agent must learn to identify and leverage patterns to achieve its objectives. (v) They can help the agent to better handle environments with changing dynamics by enabling it to learn from past experiences that may no longer be directly relevant to the current state of the environment.

\setlength{\textfloatsep}{6pt}
\begin{figure}[t]
    \scriptsize
    \centering
    \includegraphics[width=0.42\textwidth]{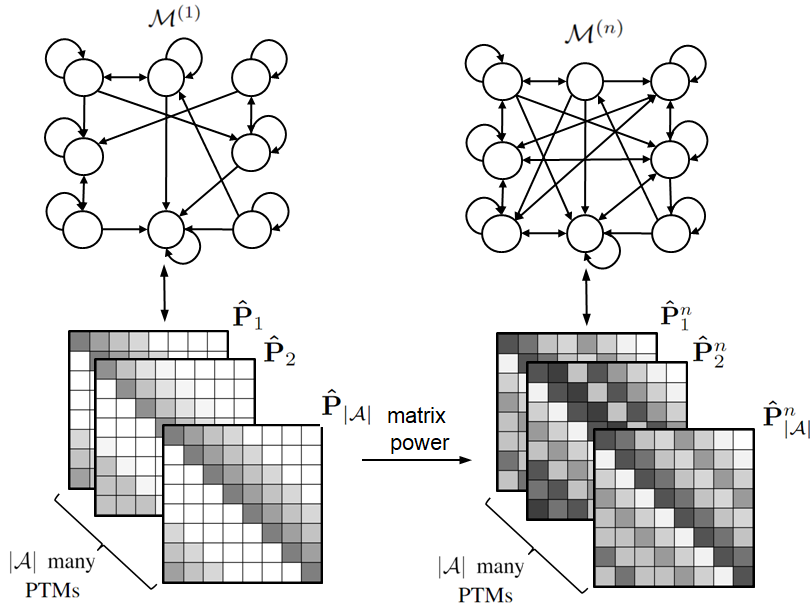}
    \caption{The relationship between $\mathcal{M}^{(1)}$ and $\mathcal{M}^{(n)}$.}
    \label{Fig:relation_between_M1_and_Mn}
\end{figure} 

The process of sampling, estimation, and constructing multiple Markovian environments is explained in Algorithm \ref{Algorithm:sampling_creating_multiple_ptts}. The inputs are the original Markovian environment, denoted by $\mathcal{M}^{(1)}$, from which the sampling process is performed, the trajectory length ($l$), the number of times each different state transition $(s \rightarrow s')$ in $\mathcal{M}^{(1)}$ must be experienced ($v$), and the total number of Markovian environments ($K$). The outputs are the $K-1$ synthetic Markovian environments (SME), denoted by $\mathcal{M}^{(n)}$ for $n = 2,3,...,K$. The underlying PTT, $\mathbf{P}$, is initially unknown as per the model-free assumption and needs to be estimated to create the PTTs of the multiple environments. Let $\mathbf{\hat{P}}$ denote the estimated PTT, and $\mathbf{\hat{P}}_a$ denote the PTM for the action $a$ in $\mathbf{\hat{P}}$ for each $a \in \mathcal{A}$. In line 1, we initialize each element of $\mathbf{\hat{P}}_a$ with $\frac{1}{|\mathcal{S}|}$ so that it is a valid PTM for each $a \in \mathcal{A}$. We keep sampling from $\mathcal{M}^{(1)}$ and updating the elements of $\mathbf{\hat{P}}$ until each one-step transition between different states under different actions in $\mathcal{M}^{(1)}$ is experienced at least $v$ times to ensure that $\mathbf{\hat{P}}$ is a sufficiently accurate estimate of $\mathbf{P}$ in lines 2-8. This procedure is known as \textit{sample averaging} \cite{sample_averaging_2}. We normalize $\mathbf{\hat{P}}_a$ row-wise such that the sum of each row in $\mathbf{\hat{P}}_a$ is 1 for each $a \in \mathcal{A}$ in line 9. We then create the $n$-hop PTTs using $\mathbf{\hat{P}}$ in line 11. Each different $\mathbf{\hat{P}}^n$ inherently corresponds to a different Markovian environment and is denoted by $\mathcal{M}^{(n)}$ in line 12. The relationship between the original Markovian environment $\mathcal{M}^{(1)}$ and the {synthetic Markovian environments} $\mathcal{M}^{(n)}$ ($n>1$) is given in Fig.\ref{Fig:relation_between_M1_and_Mn}.

Other approaches to estimate $\mathbf{P}$ include function approximations and approximate maximum likelihood estimation techniques \cite{ptm_approximate_dnn, ptm_estimate_em, truncated_ml_estimate}. However, these approaches have drawbacks: (i) Non-linear function approximations such as neural networks lack interpretability. (ii) They do not appear to offer significant computational advantages over sample averaging in sparse Markovian environments. (iii) They generally do not exploit the structural properties of the system. (iv) Training and parameter optimization can be computationally challenging.

The estimation quality affects the accuracy of $\mathbf{\hat{P}}^n$ (and thus $\mathcal{M}^{(n)}$) differently for each $n$. As $n$ increases, the error from imperfect sampling accumulates due to matrix multiplications. Consequently, higher-order environments may have low accuracy if sampling in the original environment is not done sufficiently. This suggests that $n$ should not be chosen very large for practical purposes, as we shall see in the numerical results.

\section{nEQL Algorithm and Analysis}
\label{sec:algorithm}

\algdef{SE}[REPEATN]{RepeatN}{End}[1]{\algorithmicrepeat\ #1 \textbf{times}}{\algorithmicend}
\begin{algorithm}[t]
\caption{n-hop Ensemble $Q$-Learning (nEQL)}
\hspace*{\algorithmicindent} \textbf{Input:} $l, v, u_t, K$, $\mathbf{Q}^{(n)}, \mathcal{M}^{(n)}, n \in \{1,2,...,K\}$ \\
\hspace*{\algorithmicindent} \textbf{Output: $\mathbf{Q}^{it}$, $\hat{\bm{\pi}}$} 
\begin{algorithmic}[1]
    \State Initialize $\mathbf{w}_0$ randomly, $\mathbf{Q}^{it}_0 \gets \mathbf{0}$, $t \gets 0$
    \While{each $(s,a)$ pair in $\mathcal{M}^{(1)}$ not visited $v$ times}
        \State choose common initial state for all $\mathcal{M}^{(n)}$ randomly
        \RepeatN{$l$}
            \For{each $n \in \{1,...,K\}$}
                \State sample $\{s,a,s',c\}$ from $\mathcal{M}^{(n)}$ and update $\mathbf{Q}^{(n)}_t$ using (\ref{Equ: Q-learning-update-rule})
                \State convert $\mathbf{Q}^{(n)}_t$ into probabilities $\mathbf{\hat{Q}}^{(n)}_t$ state-wise using the negative softmax
                \State $\mathbf{w}^{(n)}_t \gets 1 - \operatorname{AJSD}(\mathbf{\hat{Q}}^{(1)}_t \| \mathbf{\hat{Q}}^{(n)}_t)$
                \EndFor
            \State $\mathbf{w}_t \gets softmax(\mathbf{w}_t)$
            \State $\mathbf{Q}^{it}_{t+1} \gets u_t\mathbf{Q}^{it}_t+(1-u_t)\sum_{n=1}^{K}\mathbf{w}^{(n)}_t\mathbf{Q}^{(n)}_t$
            \State $t \gets t+1$
        \End
    \EndWhile
    \State $\hat{\bm{\pi}}(s) \gets \operatorname*{argmin}_{a'}\mathbf{Q}^{it}(s,a')$
\end{algorithmic}
\label{Algorithm: ensemble_link_learning}
\end{algorithm}

\begin{figure}[t]
    \centering
    \subfloat[Original QL]{{\includegraphics[width=1.84cm]{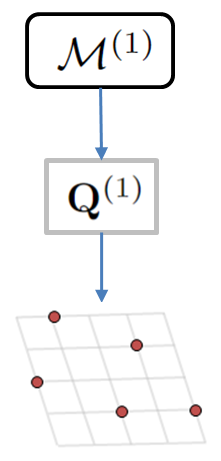}}}%
    \hspace{2pt}
    \subfloat[Ensemble QL]{{\includegraphics[width=2.92cm]{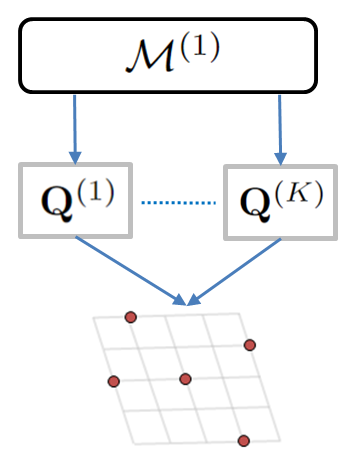}}}%
    \hspace{2pt}
    \subfloat[Proposed QL]{{\includegraphics[width=3.65cm]{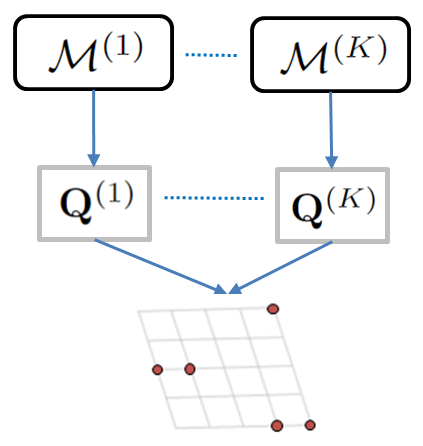}}}%
    \caption{Classification of Q-Learning (QL) algorithms based on their strategies and implementation.}
    \label{fig:different_ql_algo}
\end{figure}

In this section, we present the n-hop Ensemble Q-Learning (nEQL) algorithm (Algorithm \ref{Algorithm: ensemble_link_learning}). It is a model-free algorithm since the system dynamics, including transition probabilities and costs, are unknown. {Our approach utilizes $K-1$ {SMEs} ($\mathcal{M}^{(n)}$ for $n \in {2,...,K}$) in addition to the original Markovian environment ($\mathcal{M}^{(1)}$), resulting in a total of $K$ Markovian environments.} The high-level comparison between the original $Q$-learning algorithm, conventional ensemble $Q$-learning algorithms and proposed $Q$-learning algorithm is shown in Fig.\ref{fig:different_ql_algo}, where $\mathbf{Q}^{(n)}$ represents the $Q$-function estimator of the $Q$-learning run on $\mathcal{M}^{(n)}$ for $n\in [1,K]$.

The inputs to Algorithm \ref{Algorithm: ensemble_link_learning} consist of the trajectory length ($l$), the minimum number of visit requirement to each state-action pair ($v$), the update ratio at time $t$ ($u_t \in [0,1]$), the total number of Markovian environments ($K$), the empty $Q$-tables for $K$ different environments ($\mathbf{Q}^{(n)}$ for $n \in \{1,2,...,K\}$), and $K$ different Markovian environments ($\mathcal{M}^{(n)}$ for $n \in \{1,2,...,K\}$) since Algorithm \ref{Algorithm: ensemble_link_learning} requires access to all Markovian environments. Let $\mathbf{w}_t$ be the weight vector of size $K$ at time $t$ (with $\textbf{w}^{(n)}_t$ being the $n^{th}$ element of $\textbf{w}_t$). The weight vector at $t=0$ ($\mathbf{w}_0$) is initialized randomly to break the symmetry in line 1 ($\textit{i.e.}$ each element is chosen randomly from [0,1], and the vector is softmax-normalized so that $\sum_{n=1}^K \textbf{w}^{(n)}_0 = 1$). The iterations continue until each state-action pair in $\mathcal{M}^{(1)}$ is visited at least $v$ times (in line 2) to ensure that different state-action dynamics are sufficiently represented. 

At the end of each trajectory ($\textit{i.e.}$ every $l$ time step), all $K$ Markovian environments are reset, and a common initial state is assigned randomly from $\{1,2,...,|\mathcal{S}|\}$, as indicated in line 3. {In line 6, independent samples are collected from each different Markovian environment, and corresponding $Q$-tables are updated independently. We emphasize that given the common initial state, different actions are taken following the epsilon-greedy policy of each different environment. As a result, different next state and cost pairs are observed for different environments.} This procedure is repeated $l$ times, after which a random but common initial state is set. In line 7, the $Q$-functions are converted into probabilities per state using the negative softmax function. For example, if the $Q$-functions for a given state are [1, 1.4, 0.8, 2] (assuming four actions), we input the negative of the $Q$-functions ([-1, -1.4, -0.8, -2]) to softmax function, and compute the corresponding probabilities as [0.31, 0.21, 0.37, 0.11]. Recall that we are doing cost \textit{minimization}; thus, the smaller $Q$-function is more likely to correspond to the optimal action. In line 8, we compute the distance between the probability distributions $\mathbf{\hat{Q}}^{(1)}_t$ and {$\mathbf{\hat{Q}}^{(n)}_t$} using the \textit{averaged Jensen-Shannon divergence} ($\operatorname{AJSD}$) defined as follows:
\begin{align}
    {\operatorname{AJSD}(\mathbf{\hat{Q}}^{(1)}_t \| \mathbf{\hat{Q}}_t^{(n)}) = \frac{1}{|s|}\sum_s \operatorname{JSD}(\mathbf{\hat{Q}}^{(1)}_t(s,:) \| \mathbf{\hat{Q}}^{(n)}_t(s,:)),}
\end{align}
where $\mathbf{\hat{Q}}^{(n)}_t(s,:)$ is the probability vector of size $|\mathcal{A}|$, and $\operatorname{JSD}$ between probability distributions $p$ and $q$ is defined as \cite{jsd}:
\begin{align}
    \operatorname{JSD}(p,q) = \frac{1}{2}\Big[\operatorname{KL}\big(p\|\frac{p+q}{2}\big) + \operatorname{KL}\big(q\|\frac{p+q}{2}\big)\Big], 
\end{align}
where $\operatorname{KL}$ denotes the Kullback–Leibler divergence using base 2. Although there are several distance measures, including $l_2$ distance between $\mathbf{Q}^{(1)}_t$ and $\mathbf{Q}^{(n)}_t$ or KL divergence between $\mathbf{\hat{Q}}^{(1)}_t$ and $\mathbf{\hat{Q}}_t^{(n)}$, we employ $\operatorname{JSD}$ because (i) $\operatorname{JSD}$ is a symmetric measure in contrast to the KL divergence. (ii) JSD is a smoothed and bounded version of KL divergence (bounded to [0,1] and hence $\operatorname{AJSD}$ is also bounded to [0,1]); thus, it is robust to noise, outliers, or small perturbations in the $Q$-functions. (iii) Numerical results show that it provides superior performance to the other measures {(see \cite{talha_github})}. Herein, a larger $\mathbf{w}^{(n)}_t$ implies that the two sets of probabilities ($\mathbf{\hat{Q}}^{(1)}_t$ versus $\mathbf{\hat{Q}}^{(n)}_t$) are closer, so are the corresponding $Q$-functions ($\mathbf{Q}^{(1)}_t$ versus $\mathbf{Q}^{(n)}_t$). The vector $\mathbf{w}_t$ is softmax-normalized in line 10 and used to update {the $Q$-function output of Algorithm \ref{Algorithm: ensemble_link_learning}, $\mathbf{Q}^{it}_t$,} in line 11. When updating $\mathbf{Q}^{it}_t$, previous experience are exploited by utilizing fraction $u_t$ of the $\mathbf{Q}^{it}_t$ from the previous iteration (\textit{exploitation}), while multiple Markovian environments are sampled based on their weights (\textit{exploration}), and their contributions are weighted by $1-u_t$. In the end, the estimated policy $\hat{\bm{\pi}}$ is obtained from $\mathbf{Q}^{it}$ in line 15.

The $\mathbf{Q}^{it}_t$ (\textit{iterative}) is updated adaptively using the current weight vector $\mathbf{w}_t$; hence, it captures \textit{asymmetric} information between different Markovian environments, \textit{i.e.}, how the utility of samples obtained from $\mathcal{M}^{(n)}$ may change during iterations. It is likely that $\mathcal{M}^{(1)}$ provides more useful samples at the beginning as it is the original environment, and there are not enough samples to capture the higher-order relationships versus the first-order relationships. On the other hand, $\mathcal{M}^{(n)}$ for larger $n$ contributes more as the iterations increase as the first-order relationships may not aid as much in exploration.

The weights in Algorithm \ref{Algorithm: ensemble_link_learning} ($\mathbf{w}^{(n)}_t, n \in \{1,2,...,K\}$) converge due to the fact that the distinct $Q$-functions converge to their optimal values via $Q$-learning \cite{q_learning_convergence}, and the weights are calculated based on the attendant $Q$-functions. We will also verify the convergence of the weights numerically. 

We emphasize that there exists a potentially \textbf{distinct} optimal policy for each $K$ different Markovian environment (i.e. $\pi^*_1, \pi^*_2, ..., \pi^*_K$) corresponding to different $Q$-functions ($\mathbf{Q}^{(1)},\mathbf{Q}^{(2)},...,\mathbf{Q}^{(K)}$). Algorithm \ref{Algorithm: ensemble_link_learning} also yields an ensemble policy $(\hat{\pi})$ corresponding to the $Q$-function output $\mathbf{Q}^{it}$. We will demonstrate that $\mathbf{Q}^{it}$ converges to the optimal $Q$-functions of the original environment $({\mathbf{Q}^{(1)}}^*)$ in the mean-square sense, and thus the ensemble policy $(\hat{\pi})$ also converges to the optimal policy of the original environment $\pi^*_1$. This implies that we can obtain the optimal policy $\pi^*_1$, which is the \textbf{ultimate} goal, using our proposed algorithm with significantly lower complexity.

Our proposed algorithm combines features of online and offline RL methods. Initially, we construct the PTT of the original environment ($\hat{\mathbf{P}}$) by sampling from the original environment until $\hat{\mathbf{P}}$ is a sufficiently accurate estimate of $\mathbf{P}$. Then, we update the corresponding $Q$-functions ($\mathbf{Q}^{(1)}$) by continuously interacting with the original environment in real-time, which is the \textit{online} part of our approach. The PTTs of the multiple SMEs ($\hat{\mathbf{P}}^n$) are constructed using $\hat{\mathbf{P}}$, and the corresponding $Q$-functions ($\mathbf{Q}^{(n)}$) are updated by collecting synthetic samples from the $n^{th}$ environment, which is the \textit{offline} part. We emphasize that the estimated PTT ($\hat{\mathbf{P}}$) and PTTs of multiple SMEs ($\hat{\mathbf{P}}^n$) are constructed \textbf{only once} and not further updated using the newly collected samples in real-time, which is computationally expensive because of the need to perform matrix multiplications to construct $\hat{\mathbf{P}}^n$ and normalize $\hat{\mathbf{P}}$ and $\hat{\mathbf{P}}^n$ at each iteration. Updating $\hat{\mathbf{P}}$ beyond a certain point also results in minimal accuracy improvements, as illustrated in Fig.7. It is also important to note that our approach is \textbf{different} from hybrid RL {\cite{hybrid_rl, hybrid_rl_2, hybrid_rl_3}, where the agent generally has access to an offline dataset and subsequently collects new data through interacting with the environment.} On the other hand, our algorithm  \textit{continuously updates} the $Q$-functions in real-time while also leveraging pre-collected data to produce more accurate and stable $Q$-functions with low complexity.

\subsection{Theoretical analysis}
\label{subsec:theory}
In this section, we provide several theoretical results for Algorithm \ref{Algorithm: ensemble_link_learning}. Assume the $Q$-function {\bf errors} of the $n^{th}$ environment follows an arbitrary distribution $D_n$ with zero mean and finite variance as follows: \footnote{We observe that one can construct {\bf small} state-space examples that do not adhere to this assumption; however, for the large scale examples we consider, numerical results suggest that the assumption is valid {(see Fig.\ref{fig:Q_func_error_dist} and Fig.\ref{Fig:normal_fit})}.}
\begin{align}
    \mathcal{X}^{(n)}_t(s,a) \myeq \mathbf{Q}_{t}^{(n)}(s,a) - \mathbf{Q}^{*}(s,a)  \sim D_n\Big(0,\frac{\lambda_n^2}{3}\Big), \label{Equ: distribution_assumption}
\end{align}
for all $(s,a)$ and $n$ with $\lambda_n > 0$ where $\mathbf{Q}^{*}$ is the optimal $Q$-functions of the original Markovian environment. Prior work has considered the distribution $D_n$ to be uniform, non-uniform, or normal {for the $n=1$ case} \cite{uniform_assump_1, deep_double_q, gaussian_approx_1, randomized_double_q}. Herein, we make no assumptions on $D_n$. Simulations verify that the true distributions $D_n$ are, in fact, very close to the normal distributions with zero-mean and finite variance for all $n$ {(see Fig.\ref{fig:Q_func_error_dist} and Fig.\ref{Fig:normal_fit})}.

Let $\mathbb{E}$ and $\mathbb{V}$ be the expectation and variance operators, $\lambda = \max\limits_{n \in \{1,2,...,K\}}\lambda_n$, and $\mathcal{E}_t(s,a) \myeq \mathbf{Q}_{t}^{it}(s,a) - \mathbf{Q}^{*}(s,a)$. 
\begin{proposition}\label{proposition-1}\normalfont {Let $u_t$ be a constant: $u_t=u$}. Under Assumption (\ref{Equ: distribution_assumption}), Algorithm \ref{Algorithm: ensemble_link_learning} produces unbiased $Q$-functions in the limit $\textit{i.e.}$ $\lim\limits_{t\rightarrow \infty}\mathbb{E}[\mathcal{E}_t(s,a)] = 0$. If the $Q$-function errors of a given environment at different times are independent {$\textit{i.e.}$ $\mathcal{X}^{(n)}_{t_1}(s,a) \perp \mathcal{X}^{(n)}_{t_2}(s,a)$ for all $s,a,n$, $t_1 \neq t_2$}, the error variance in the limit can be upper bounded as: $\lim\limits_{t\rightarrow \infty}\mathbb{V}[\mathcal{E}_t(s,a)] \leq \frac{(1-u)}{(1+u)}\lambda^2$. (see Appendix \ref{Appendix: proposition_1})
\end{proposition}

This proposition shows that under the assumption (\ref{Equ: distribution_assumption}), $\mathbf{Q}_{t}^{it}$ is an unbiased estimator of $\mathbf{Q}^{*}$ in the limit, {and the upper bound on the error variance can be controlled by the parameters $u$ and $\lambda$.} Herein, a larger $\lambda$ implies a higher uncertainty in the $Q$-function errors, which makes the upper bound on the variance looser. {On the other hand, when the algorithm converges ($t \rightarrow \infty$), a larger $u$ leads to less reliance on SMEs, reducing the uncertainty arising from multiple environments and yielding a tighter upper bound.}

The zero-mean assumption in (\ref{Equ: distribution_assumption}) is only employed to simplify the analysis of Proposition \ref{proposition-1}. Nevertheless, this assumption can be relaxed as follows:
\begin{corollary}\label{corollary_relaxation}\normalfont
Proposition \ref{proposition-1} is valid under the assumption that the $Q$-function errors follow arbitrary distributions, that is:
\begin{align}
    \mathcal{X}^{(n)}_t(s,a) \myeq \mathbf{Q}_{t}^{(n)}(s,a) - \mathbf{Q}^{*}(s,a)  \sim D_n\Big(\mu_n,\frac{\lambda_n^2}{3}\Big), \label{Equ: simplified_distribution_assumption}
\end{align}
if the weighted combination of the means at different times is $0$, i.e., $\sum_{n=1}^K\mathbf{w}_t^{(n)}\mu_n = 0$, where $\mu_n$ is the mean of the distribution $D_n$. (see Appendix \ref{Appendix: corollary_1})
\end{corollary}

This assumption is less restrictive than (\ref{Equ: distribution_assumption}) since it only requires the weighted convex combination of means to be zero, allowing different means to be non-zero. This result can be numerically validated by Fig.\ref{fig:Q_func_error_dist} and Fig.\ref{Fig:normal_fit}.

We emphasize that there is estimation bias between $\mathbf{Q}^{(1)}$ and $\mathbf{Q}^{(n)}$ for $n>1$ since these $Q$-functions are obtained from different environments, and this is reflected by the non-zero means in  Equation (\ref{Equ: simplified_distribution_assumption}); however, the sign of estimation error (i.e., whether it is overestimation or underestimation) depends on $n$. As will be seen later, Algorithm 2 produces unbiased output ($\mathbf{Q}^{it}$). This follows as (i) Algorithm 2 uses a weighted combination of different $Q$-functions to update $\mathbf{Q}^{it}$ and thus the effect of underestimation and overestimation cancel each other, and (ii) a small weight is assigned to the $Q$-function with high bias, minimizing the impact of individual biases.

\begin{corollary}\label{corollary0}\normalfont
If we remove the independence assumption in Proposition \ref{proposition-1}, the upper bound on the error variance in the limit can be updated as: $\lim\limits_{t\rightarrow \infty}\mathbb{V}[\mathcal{E}_t(s,a)] \leq \frac{2\lambda^2}{(1+u)^2}+\frac{(1-u)}{(1+u)}\lambda^2$. {(see \cite{talha_github})}
\end{corollary}

{The relaxation of the independence assumption introduces a bias term to the upper bound estimation, resulting in a looser bound than the one in Proposition \ref{proposition-1}. Nevertheless, a smaller $\lambda$ or a larger $u$ tightens the bound as in Proposition \ref{proposition-1}.}

\begin{corollary}\label{corollary}\normalfont
If we use the form $u_t = 1 - e^{\frac{-t}{c_4}}$ with $c_4 > 0$ and the independence assumption in Proposition \ref{proposition-1}, the error variance converges to zero: $\lim\limits_{t\rightarrow \infty}\mathbb{V}[\mathcal{E}_t(s,a)] = 0$. 
\end{corollary}

We choose the parameter $u_t$ such that $u_t \xrightarrow{t \rightarrow \infty} 1$ as $u$ should be small initially to explore multiple environments in the beginning (exploration) and should increase to utilize previously obtained samples with time (exploitation $\textit{i.e.}$ less reliance on the synthetic environments as we learn $\mathbf{Q}_t^{it}$ better). Herein, using multiple Markovian environments accelerates the convergence of $\mathbf{Q}_t^{it}$ towards $\mathbf{Q}^*$, while adjusting $u_t$ pushes $\mathbf{Q}_t^{it}$ in the desired direction. Consequently, Algorithm \ref{Algorithm: ensemble_link_learning} converges and yields the optimal $Q$-functions \textit{in the mean-square sense}, which can be shown by combining the results of Proposition \ref{proposition-1} and Corollary \ref{corollary}. We will numerically verify that the independence assumption is \textbf{almost always} satisfied. The parameter $c_4$ adjusts the decay rate of $u_t$, which is crucial to tune the amount of exploration. In particular, a larger $c_4$ implies a slower decay rate, which is needed for larger networks, where more exploration is necessary. This result aligns with Proposition \ref{proposition-1} and Corollary \ref{corollary0}, as larger values of $u$ lead to tighter upper bounds on the error variance and $u = 1$ makes the upper bounds zero.

The structure of $u_t$ ($u_t \xrightarrow{t \rightarrow \infty} 1$) prevents error accumulation in $\mathbf{Q}^{it}$ due to the potential lack of knowledge about $\mathbf{Q}^{(1)}$, especially in the initial stages. Even if weights at time $t$ ($\mathbf{w}_{t}$) are incorrectly assigned, potentially leading to errors in computing $\mathbf{Q}^{it}_{t+1}$, the impact of this contribution to the final $Q$-function estimate is outweighed by that of $\mathbf{Q}^{it}_{t+2}$ due to scaling by $u_t$, which increases over time ($u_t < u_{t+1}$). Herein, $\mathbf{Q}^{it}_{t+2}$ is calculated using updated weights ($\mathbf{w}_{t+1}$), which are computed based on a more accurate estimate of $\mathbf{Q}^{(1)}$.

The convergence proof of the algorithm in our prior work \cite{talha_icassp} can be adapted to show the convergence of the Algorithm \ref{Algorithm: ensemble_link_learning} deterministically without any distribution or independence assumption on the $Q$-function errors.

\begin{proposition}\label{proposition-2}\normalfont
{The upper bound on the error variance decreases with the number of Markovian environments $K$ as $\mathbb{V}[\mathcal{E}_t(s,a)] \leq \frac{c(\lambda, u)}{K}$ for all $t$, where $c(\lambda, u)$ is a constant of $K$. (see Appendix \ref{Appendix: proposition_2})}
\end{proposition}

The proposition shows that increasing the number of Markovian environments ($Q$-learning algorithms running on different environments simultaneously) reduces the upper bound on variance. This aligns with the primary objective of ensemble algorithms. Unlike the bound in Proposition \ref{proposition-1}, this upper bound explicitly depends on $K$, is valid without any assumptions on independence or structure of $u_t$, and holds for all $t$. The decay rate of the upper bound is determined by the function $c(\lambda, u)$, which incorporates both $\lambda$ and $u$. {Furthermore, numerical results will show that increasing $K$ may not always yield an increasing reduction in the upper bound, $\textit{i.e.}$ there is a diminishing return of increasing $K$.}

\begin{proposition}\label{proposition-3}\normalfont
Let the output policy of Algorithm \ref{Algorithm: ensemble_link_learning} be $\hat{\pi}$, and the $Q$-functions of the $n^{th}$ environment under the policy $\hat{\pi}$ be $\mathbf{Q}^{(n)}_{\hat{\pi}}$. Then, Algorithm \ref{Algorithm: ensemble_link_learning} produces $Q$-functions on different environments that satisfy the following upper bound:
\begin{align}
    \|\mathbf{Q}^{(1)}_{\hat{\pi}} - \mathbf{Q}^{(n)}_{\hat{\pi}}\| < \frac{\gamma}{1-\gamma^n}\frac{1-\gamma^{n-1}}{1-\gamma}\|\mathbf{c}_{\hat{\pi}}\|,\label{Equ: temp_proposition}
\end{align}
where $n > 1$, the norm $\|\cdot\|$ is the $l_2$ norm, and $\mathbf{c}_{\hat{\pi}}$ is the cost vector under the policy $\hat{\pi}$. (see Appendix \ref{Appendix: proposition_3})
\end{proposition}

This proposition shows that the $Q$-functions of different environments under the same policy are closely related as a result of the structural relationship between different Markovian environments. {As $n \rightarrow \infty$, the upper bound primarily depends on the cost function, eliminating the influence of the learning parameter $\gamma$. This suggests $n$ should not be chosen very large for practical purposes.} Note that this behavior should not imply the monotonicity of the $Q$-functions as a function of $n$, as will be seen later, $\textit{i.e.}$ $\|\mathbf{Q}^{(1)}_{\hat{\pi}} - \mathbf{Q}^{(n)}_{\hat{\pi}}\|$ is a non-monotonic function of $n$. It is also worth emphasizing that this result holds without any assumptions on the independence or structure of $u_t$.

\begin{proposition}\label{proposition-4}\normalfont
The $\mathbf{Q}^{(n)}_{\hat{\pi}}$ satisfy the following (partial) ordering when $\gamma \rightarrow 1$ ($\textit{i.e.}$ when the underlying discounted MDP starts to resemble an undiscounted MDP): (see Appendix \ref{Appendix: proposition_4})
\begin{align}
    \mathbf{Q}^{(n)}_{\hat{\pi}}(s) &\geq \mathbf{Q}^{(2n)}_{\hat{\pi}}(s) \geq \mathbf{Q}^{(4n)}_{\hat{\pi}}(s) \geq \mathbf{Q}^{(8n)}_{\hat{\pi}}(s) ..., 
\end{align}
for all $s$, where $\mathbf{Q}^{(1)}_{\hat{\pi}}(s)$ is the largest and $n$ is an odd number.
\end{proposition}

This proposition enables assessing the utility of various Markovian environments by ordering the $Q$-functions and facilitates determining the most useful (informative) environments to be used in Algorithm \ref{Algorithm: ensemble_link_learning}.

Assume we want to use $K$ = 4 environments. A challenge is to determine which combination of four environments to include in our algorithm, for example: $\{1^{st},2^{nd},3^{rd},4^{th}\}$ vs $\{1^{st},2^{nd},3^{rd},5^{th}\}$, etc. Herein, the notation $n^{th}$ simply refers to the matrix power used to create Markovian environments. It is important to note that $n$ can be greater than $K$, (i.e., $n$ is not limited to $[1, K]$). Thus, in this example, using the $5^{th}$ environment does not necessarily require using the first five environments as the $4^{th}$ environment is not used. Herein, we want to collect diverse $n$-hop information from different environments while avoiding environments that do not contribute much to achieving the optimal solution. To this end, we assess the similarity of $\mathbf{Q}^{(n)}$ (for $n > 1$) to $\mathbf{Q}^{(1)}$ based on the given orderings. $\mathbf{Q}^{(1)}$ and $\mathbf{Q}^{(2)}$ are the most similar, making it a logical choice to include both environments. The $3^{rd}$ environment is added to explore its potential usefulness, given the uncertain similarity between $\mathbf{Q}^{(3)}$ and $\mathbf{Q}^{(2)}$. In contrast, $\mathbf{Q}^{(4)}$ is less similar to $\mathbf{Q}^{(1)}$ than $\mathbf{Q}^{(2)}$; thus, we do not include the $4^{th}$ environment, while we can potentially explore more useful environments. Due to the same reason, we also want to explore the $5^{th}$ environment. Hence, a reasonable selection is to use $1^{st},2^{nd},3^{rd}$ and $5^{th}$ environments. We emphasize that we use $K=4$ environments, but one of the environments is constructed using the $5^{th}$ matrix power, which is greater than $K$. While this approach may not always be optimal, it offers a pragmatic strategy when prior information is lacking.

We emphasize that this result is valid without any independence or structural assumption on $u_t$. Moreover, this result is particularly useful for network settings when long-term planning is more important, for example, when the future costs are more important than the immediate costs or taking some actions may not minimize the immediate rewards, but they will be more beneficial in the long run.

\section{Numerical Results}
\label{sec:numerical_results}

In this section, we consider a variety of performance metrics to assess the accuracy and complexity performance of Algorithm \ref{Algorithm:sampling_creating_multiple_ptts} and \ref{Algorithm: ensemble_link_learning} across different network models.

\subsection{Network models}
\label{subsec:wireless_network_models}

We consider four different network models, which differ in their design, complexity, and implementation. {(See \cite{talha_github} for further details)}

\begin{figure}[t]
    \centering
    \subfloat[SISO network model]{{\includegraphics[width=5.1cm]{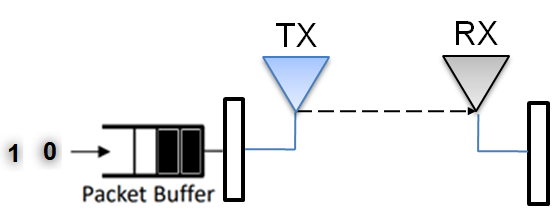}}\label{fig:siso_network}}%
    \vspace{10pt}
    \subfloat[MISO network energy harvesting model with relays]{{\includegraphics[width=7.2cm]{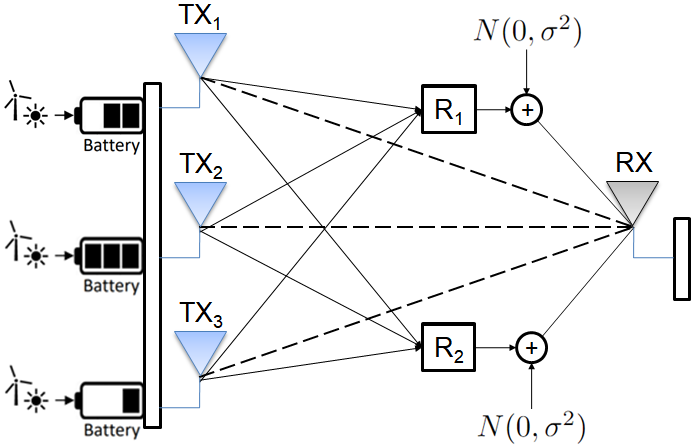}}\label{fig:miso_network}}%
    \caption{Examples of wireless network models.}
\end{figure}

\subsubsection{Randomized graphs} We consider the Erdős-Rényi (ER) random-directed graph model. The PTM has $|\mathcal{S}|$ nodes, and each edge is created with a probability $0.2$. We create $|\mathcal{A}|$ many PTMs and concatenate them to obtain the PTT, which is used for sampling and creating the {SMEs}. The cost function assigns a uniform $[0,1]$ random cost to each state-action pair.

\subsubsection{Cliff-walking environment} We consider the cliff-walking environment \cite{barto_sutton_rl}. The number of columns is chosen to be approximately three times the number of rows in the grid so that the number of states is equal to $|\mathcal{S}|$ (for example, 60 columns and 20 rows $\rightarrow$ 1800 states). If the agent moves to the cliff region, a unit cost is incurred. Moving to the safe grid results in a negative unit cost, while any other movement incurs a cost of 0.01.

\subsubsection{SISO wireless network model} We consider the model of \cite{libin_paper}, in which there is a single transmitter (TX) and receiver (RX) as shown in Fig.\ref{fig:siso_network}. The goal is to determine when the transmitter should \textit{transmit} data or \textit{remain silent} to minimize the sum of transmission and packet drop costs.

\subsubsection{MISO energy harvesting wireless network with Gaussian interference channels and multiple relays} We consider the model of \cite{colink_journal}. An example network with three transmitters (TX$_1$, TX$_2$, TX$_3$), a single receiver (RX), and two relays (R$_1$, R$_2$) is shown in Fig.\ref{fig:miso_network}. The goal is to determine when transmitters should \textit{directly} transmit or transmit \textit{through relays} in order to maximize the overall throughput while minimizing the sum of battery and packet drop costs for each transmitter. 

\begin{figure}[t]
    \centering
    \includegraphics[width=0.36\textwidth]{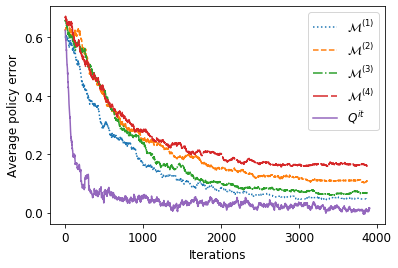}
    \caption{APE performances across different environments.}
    \label{Fig:sharp_decrease}
\end{figure} 

\subsection{Average Policy Error Results}\label{Subsec:APE}
Let $\bm{\pi^{*}}$ be the optimal policy from (\ref{Equ: optimization_eq_2}), and $\hat{\bm{\pi}}$ be the output policy of Algorithm \ref{Algorithm: ensemble_link_learning}. Since our main concern is optimal control, we define the \textit{average policy error (APE)} as follows:
\begin{align}
    APE &=\frac{1}{|\mathcal{S}|} \sum_{s=1}^{|\mathcal{S}|} \mathbf{1}\left(\bm{\pi^{*}}(s) \neq \hat{\bm{\pi}}(s)\right).
\end{align}
We analyze the performance of Algorithm \ref{Algorithm: ensemble_link_learning} over $Q$-learning in Fig.\ref{Fig:sharp_decrease}. The simulation is carried out using model-4 with network size 5000 with the following parameters: $\gamma=0.95$, $\alpha_t$=$\frac{1}{1+\frac{t}{100}}$, $u_t=1-e^{\frac{-t}{1000}}$, $v=40$, $l=10$, $K=4$. These parameters are optimized through cross-validation (see Section \ref{subsec: Hyper-parameter tuning} for further details). We choose the $epsilon$ value for each distinct $Q$-learning algorithm as follows: $\epsilon_t^{(n)} = \max((c_n)^t, 0.01)$ for $n=1,2,3,4$ with $c_1 = 0.95, c_2 = c_3 = 0.97, c_4 = 0.99$. We choose these parameters to prioritize learning from useful environments over random exploration. Faster decay rates (i.e., smaller $c_n$) are applied to environments that closely resemble the original environment, informed by Proposition \ref{proposition-4}. The curves represent APE of $Q$-learning algorithms on the original environment $\mathcal{M}^{(1)}$ and three different {SMEs} $\mathcal{M}^{(n)}$ for $n = {2,3,4}$, while Q$^{it}$ represents the APE of Algorithm \ref{Algorithm: ensemble_link_learning}. Clearly, a near-zero APE (around 0.05) can be achieved with a significantly small number of iterations (around 500). The sharp decline in the $Q^{it}$ curve at the beginning (up to the 200$^{th}$ iteration) corresponds to the \textit{exploration} stage, followed by the \textit{exploitation} stage.  Compared to any other algorithm, the exploration stage in Algorithm \ref{Algorithm: ensemble_link_learning} is fast, which shows the advantages of utilizing multiple Markovian environments to enable a deep and efficient exploration and accelerate the overall training. Notice that individual $Q$-learning algorithms can only achieve a slightly higher APE than that of Algorithm \ref{Algorithm: ensemble_link_learning} if they run for a significantly long time. We observe that the APE results are not monotonic across $n$ (APE of $Q$-learning run on $\mathcal{M}^{(2)}$ is lower than that of $\mathcal{M}^{(4)}$), which is in line with Proposition \ref{proposition-4}.

In order to provide a performance comparison, we employ several $Q$-learning algorithms, each with different objectives and strategies (number of estimators). Table \ref{Table:Q-learning-variants} provides an overview of these algorithms. Specifically, we focus on value-based model-free algorithms which follow the same strategy as Algorithm \ref{Algorithm: ensemble_link_learning} to ensure a fair comparison. We also include two algorithms using multiple models (environments): \textit{Ensemble Graph Q-Learning} (EGQL) \cite{talha_icassp} and \textit{Model-ensemble trust-region policy optimization} (TRPO) \cite{kurutach2018model}. EGQL adopts a different strategy to create multiple SMEs instead of $n$-hop systems, employs a weighting mechanism based on policies rather than $Q$-functions, and imposes several constraints and assumptions on the system model. TRPO takes a different approach by using an ensemble of deep neural networks for modeling both dynamics and policy (model-based strategy compared to our hybrid scheme). It fits this ensemble to a single real-world dataset (in contrast to our use of multiple datasets, some synthetically generated) and employs supervised learning for training, unlike this work. For further details regarding the parameter optimization of each algorithm, refer to \cite{talha_github}.

\begin{table}[t]
\centering
\scriptsize
\setlength{\tabcolsep}{2pt}
\begin{tabular}{|c|c|c|c|}
 \hline
 Algorithm & Objective & \multicolumn{2}{c|}{Strategy} \\ [0.5ex]
 \cline{3-4}
 & & Est. & Env. \\ [0.5ex]
 \hline\hline
 Simple Q (Q) \cite{bertsekas_book} & - & Single & Single \\ [0.5ex]
 \hline
 Speedy Q (SQ) \cite{speedy_q} & Convergence rate & Single & Single\\ [0.5ex]
 \hline
 Double Q (DQ) \cite{double_q} & Bias & Multi & Single\\ [0.5ex]
 \hline
 MaxMin Q (MMQ) \cite{maxmin-q} & Bias \& variance & Multi & Single\\ [0.5ex]
 \hline
 Ensemble Bootst. Q (EBQ) \cite{ensemble_bootstrap_q} & Bias & Multi & Single\\ [0.5ex]
 \hline
 Averaged DQN (ADQN) \cite{averaged_dqn} & Stability, Variance & Multi & Single\\ [0.5ex]
\hline
 Model Ensemble (TRPO) \cite{kurutach2018model} & Sample Complexity, Stability & Multi & Multi\\ [0.5ex]
 \hline
 Ensemble Graph Q (EGQL) \cite{talha_icassp} & Variance, Learning speed & Multi & Multi\\ [0.5ex]
 \hline
\textbf{n-hop Ensemble Q (nEQL)} & \textbf{Variance, Learning speed} & \textbf{Multi} & \textbf{Multi}\\ [0.5ex]
\hline
\end{tabular}
\caption{$Q$-learning algorithm and variants.}
\label{Table:Q-learning-variants}
\end{table}

\begin{figure}[t]
    \centering
    \subfloat[APE for model-1\label{fig:ape_model1}]
    {{\includegraphics[width=4.45cm]{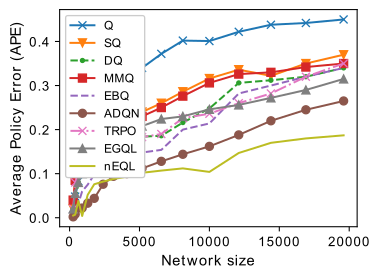}}}
    \subfloat[APE for model-2\label{fig:ape_model2} ]{{\includegraphics[width=4.45cm]{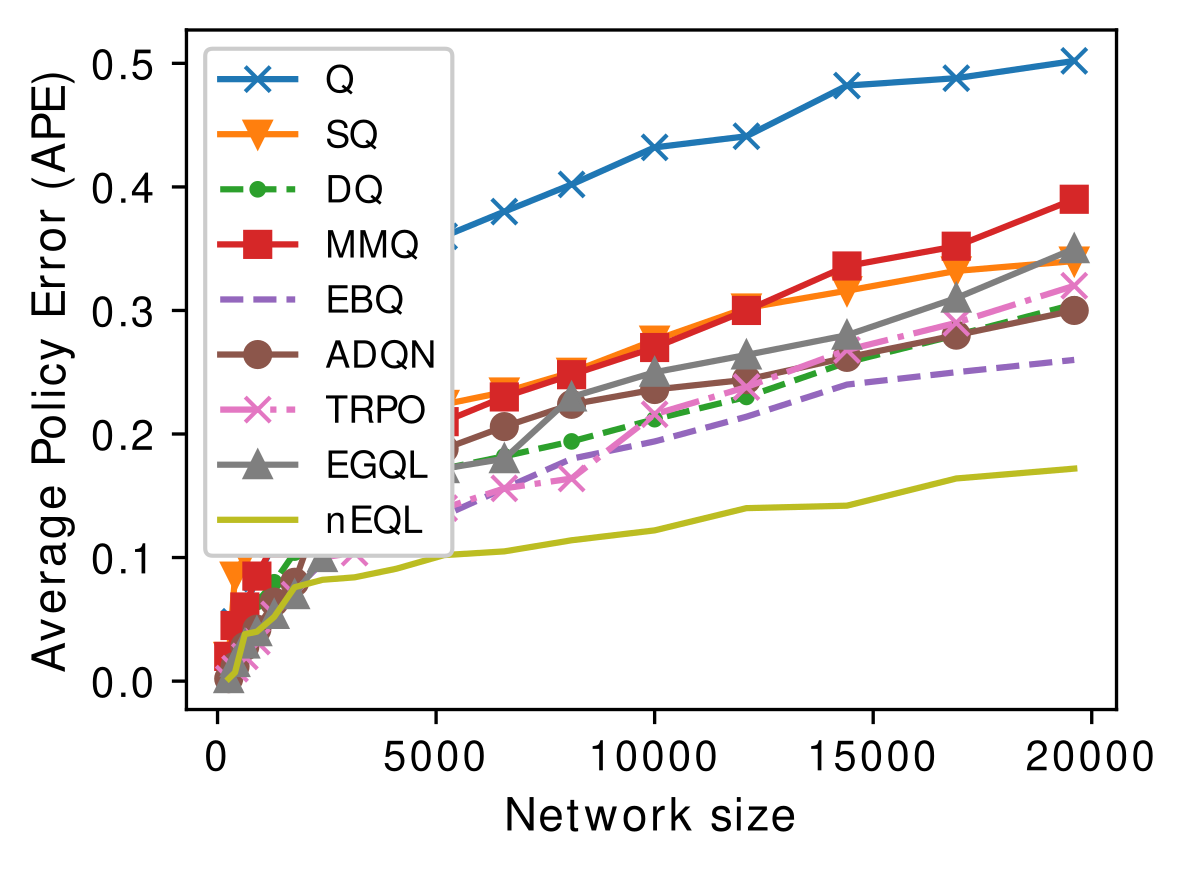}}}

    \subfloat[APE for model-3 \label{fig:ape_model3} ]{{\includegraphics[width=4.45cm]{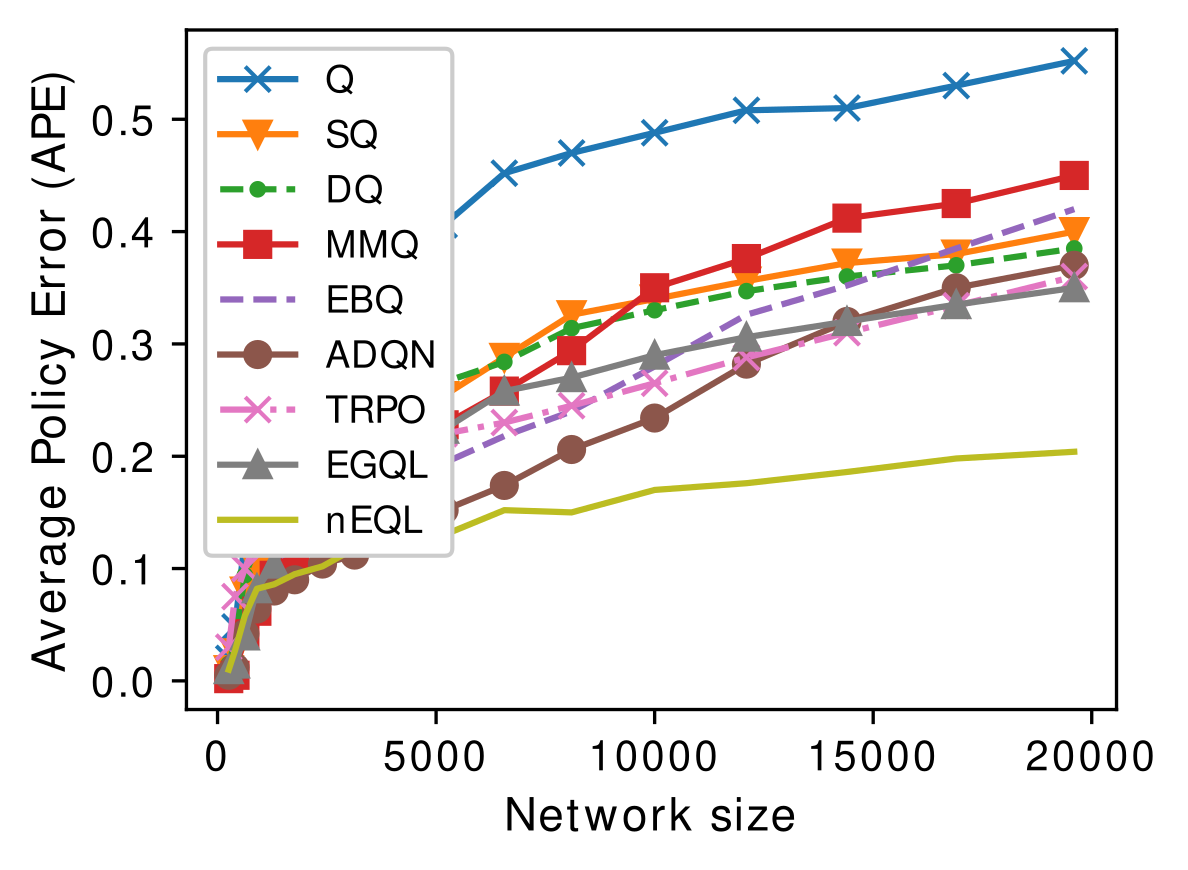}}}
    \subfloat[APE for model-4 \label{fig:ape_model4}]
    {{\includegraphics[width=4.45cm]{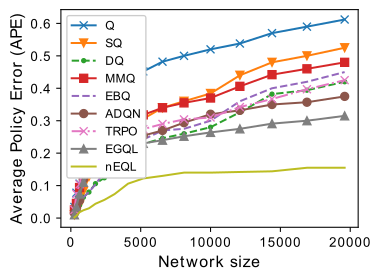}}}
    \caption{APE results across different network models.}
\end{figure}

The APE of different algorithms across network size and different models are given in Fig.\ref{fig:ape_model1}-\ref{fig:ape_model4}. Overall, our proposed algorithm consistently achieves a lower APE compared to other algorithms: 30\% less for model-1, 35\% less for model-2, 45\% less for model-3, and 55\% less for model-4. The APE gains become more clear for larger networks as using multiple Markovian environments enables deep and efficient exploration by combining the $n$-hop relationships between states into a single estimate, and the weighting mechanism based on $\operatorname{JSD}$ enables the algorithm fully exploit the most useful environments during training by assigning higher weights. 

The proposed algorithm demonstrates significant APE gains, particularly with model-3 and model-4, showing its effectiveness and practicality across real-world networks. In general, ADQN, EBQ, and EGQL produce the lowest APE among other algorithms; however, they have inferior performance compared to nEQL because neural network-based algorithms (ADQN and TRPO) do not leverage the structural properties of multiple Markovian environments. The performance of EBQ shows the advantages of ensemble algorithms. However, it does not employ a weighting mechanism when calculating the ensemble but uses simple averaging; thus, it performs worse than nEQL, which employs an adaptive weighting mechanism. The performance of EGQL underscores the benefits of using multiple environments (i.e., model ensemble). Nevertheless, it relies on a less accurate and less robust weighting mechanism, does not make use of the simplicity of the $n$-hop systems, and suffers from the constraints on the system model. We also observe that as the model complexity increases (from model-1 to model-4), the APE of all algorithms increases. The proposed algorithm, however, yields the least increase in APE, making it the most accurate algorithm. The APE order of the other algorithms remains consistent across different models.

The network models considered herein can be well-modeled with discrete state-spaces with $|\mathcal{S}|$ < 20000 (as in \cite{talha_eusipco, talha_icassp, colink_journal, libin_paper}). Numerical simulations show that the performance between the proposed algorithm (tabular-based) and ADQN  (neural-network-based) varies as a function of the network size. While there is no clear performance difference for small networks ($|\mathcal{S}|$ < 1000), the proposed algorithm offers up to 35\% less APE with 20\% less runtime for modest-sized networks (1000 < $|\mathcal{S}|$ < 10000). For large networks (10000 < $|\mathcal{S}|$ < 20000), the advantages of our algorithm become more apparent (55\% less APE with 50\% less runtime) because of (i) the challenging state-space design for ADQN, (ii) the increased overfitting risk in sparse state spaces for ADQN in data-limited regions, and (iii) the simple averaging mechanism in ADQN that does not capture the changes in the distribution of the $Q$-functions over time.

We observe that our algorithm inherits several properties from the traditional $Q$-learning algorithm and hence is particularly tailored to finite, but large discrete state-action spaces. For networks with continuous state-spaces, the tabular nature of our algorithm may pose limitations. To this end, replacing tabular $Q$-learning with deep $Q$-networks is worth exploring.

\subsection{Computational Complexity Results}\label{Subsec:runtime_complexity}
The algorithm \ref{Algorithm: ensemble_link_learning} can be shown to have the following average time-complexity $O\left(\frac{|\mathcal{S}||\mathcal{A}| v}{K}f(l,\epsilon)\right)$, where $f$ is some non-monotonic function of $l$ and $\epsilon$. {The derivation closely follows that presented in \cite{talha_eusipco}}. The runtime complexity increases with the network size ($|\mathcal{S}|$ and $|\mathcal{A}|$) as well as the number of visit requirement to each state-action pair ($v$). On the other hand, the non-monotonicity of $f$ in $l,\epsilon$ implies that there are optimal values for $l, \epsilon$; thus, parameter-tuning is required. The complexity is also inversely proportional to the number of Markovian environments ($K$), which may seem counter-intuitive. However, the number of samples that need to be collected from each Markovian environment decreases with $K$, leading to an overall reduction in the training runtime complexity.

\begin{figure}[t]
    \centering
    \includegraphics[width=0.3\textwidth]{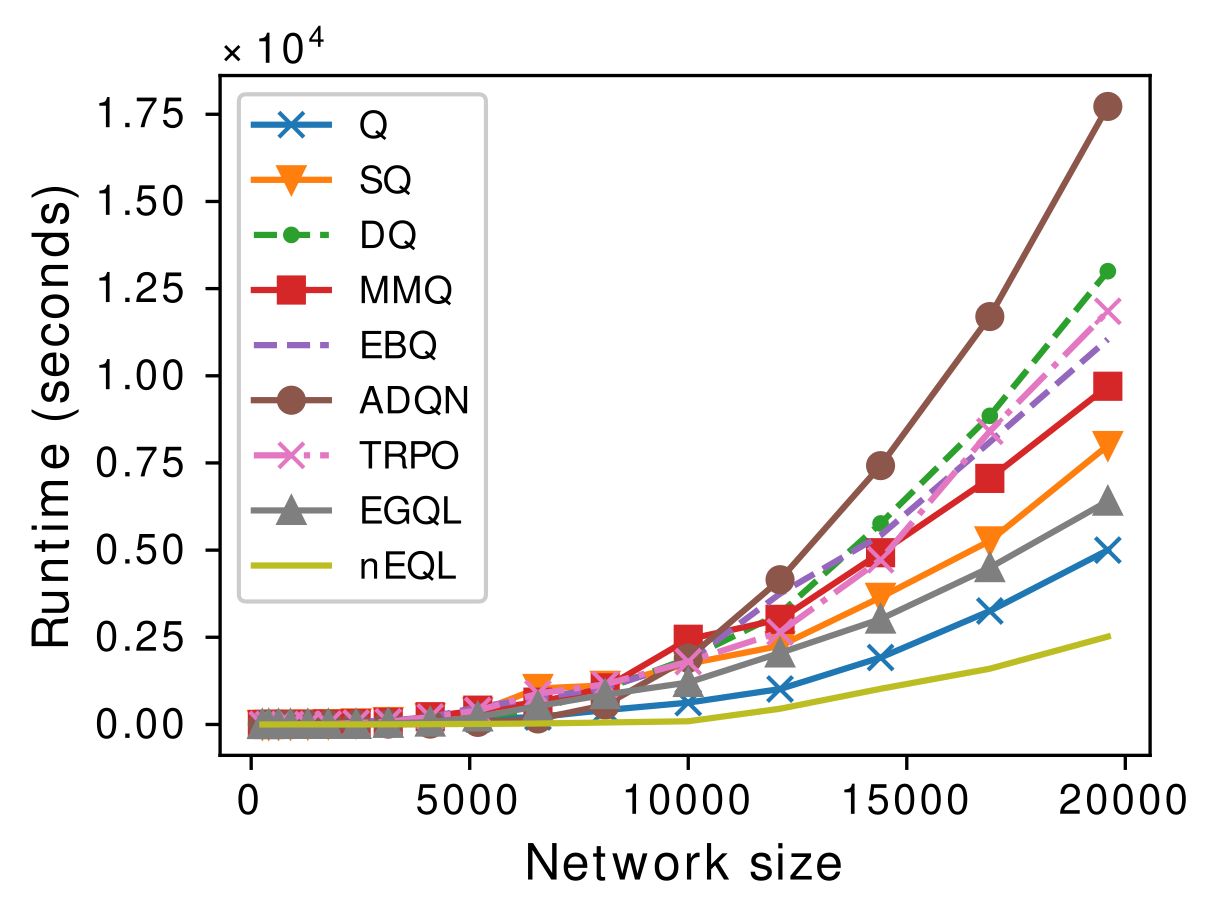}
    \caption{Runtimes for different algorithms.}
    \label{Fig:runtime}
\end{figure} 

The runtime of nEQL includes time spent on sampling (visiting each state-action pair at least $v$ times), constructing multiple {SMEs} $K-1$ times using matrix power operations, and the time until convergence of $Q$-learning algorithms on different Markovian environments. The overall runtime is determined by the slowest $Q$-learning algorithm, as they are run in parallel. Other algorithms' runtime results represent the time until each algorithm converges. This runtime measure also serves as a measure of \textit{computational power complexity}. Across fixed network sizes and four different models, the runtime of algorithms is comparable. Therefore, a single runtime result is presented in Fig.\ref{Fig:runtime}. The proposed algorithm achieves 50\% less runtime than the other algorithms across large networks, which can be attributed to several reasons: (i) Utilizing multiple Markovian environments reduces the need for long trajectories to capture distant node relationships. (ii) The amount of exploration is improved due to the running of multiple Markovian environments simultaneously. (iii) The algorithm can be terminated early with minimal APE as shown in Fig.\ref{Fig:sharp_decrease}. (iv) The average runtime complexity decreases with $K$ hyperbolically. (v) Employing different epsilon-greedy policies allows us to capture the distinct characteristics of different Markovian environments. The complexity reduction is independent of the network model or underlying assumptions, making the proposed algorithm an efficient approach for learning various complex environments.

The original $Q$-learning has a relatively lower runtime than the other benchmarks as it is a straightforward algorithm. SQ aims to improve the learning (and convergence) rate; therefore, it also produces a relatively lower runtime. EGQL also uses a strategy similar to the current approach using multiple models, which makes it relatively faster than other benchmarks. On the other hand, EBQ has multiple estimators, while ADQN and TRPO suffer from the need for long training periods, which becomes computationally intractable for large networks. It is worth noting that as the model complexity increases (e.g., Q $\rightarrow$ NQ $\rightarrow$ ADQN), the corresponding APE generally decreases, but the runtime complexity increases. However, nEQL achieves a small APE with a small runtime, overcoming the performance-complexity trade-off.

To solve (\ref{Equ: optimization_eq}) and (\ref{Equ: optimization_eq_2}), one could consider using dynamic programming methods such as value iteration instead of $Q$-learning, given that we estimate the PTTs of multiple Markovian environments using $\mathbf{\hat{P}}^n$. We herein follow a similar iterative procedure to Algorithm \ref{Algorithm: ensemble_link_learning}. While we keep sampling to update $\mathbf{\hat{P}}$ continuously, we construct the PTTs of multiple {SMEs} using the current estimate of $\mathbf{\hat{P}}$ every 10 iterations, and obtain the value functions $\mathbf{v}^{(n)}_t$ for $n \in \{1,2,...,K\}$ using the value iteration algorithm. We then assign the weights $\mathbf{w}^{(n)}_t$ as:
\begin{align}
    \mathbf{w}^{(n)}_t &\gets -\|\mathbf{v}^{(1)}_t-\mathbf{v}^{(n)}_t\|_2, \hspace{4pt} \forall n\\
    \mathbf{w}_t &\gets \mbox{softmax}(\mathbf{w}_t),
\end{align}

We finally update the value function of the proposed algorithm $\mathbf{v}^{it}_t$ similar to Algorithm \ref{Algorithm: ensemble_link_learning} using $u_t$ (with the same form as in Corollary \ref{corollary}). We carry out these operations iteratively $k$ times, where $k$ is sufficiently large to ensure the convergence. This approach has runtime-complexity $O(k|\mathcal{S}|^2|\mathcal{A}|)$ since we carry out value iteration $k$ times, which has a complexity of $O(|\mathcal{S}|^2|\mathcal{A}|)$. In our simulations, with the settings in Section \ref{Subsec:APE}, this strategy incurs 50\% more runtime complexity and 60\% more APE across large networks, making it inferior to our proposed algorithm.

In order to show the computational advantages of Proposition 4 in determining the optimal set of Markovian environments to be employed in Algorithm \ref{Algorithm: ensemble_link_learning}, we choose $K=4$ environments over the set of orders $n = \{1,2,3,4,5,6\}$ by (i) using partial ordering from Proposition 4 and (ii) trying every possible  distinct combination of four Markovian environments and choosing the one that minimizes APE (i.e. exhaustive search). The same simulation settings in Section \ref{Subsec:APE} are employed. Numerical results show that Proposition 4 gives the optimal set of Markovian environments with 35\% less runtime complexity than exhaustive search. The complexity reduction further increases with the network size and $K$, which implies that Proposition 4 is particularly useful for large networks.

\begin{figure}[t]
    \centering
    \includegraphics[width=0.3\textwidth]{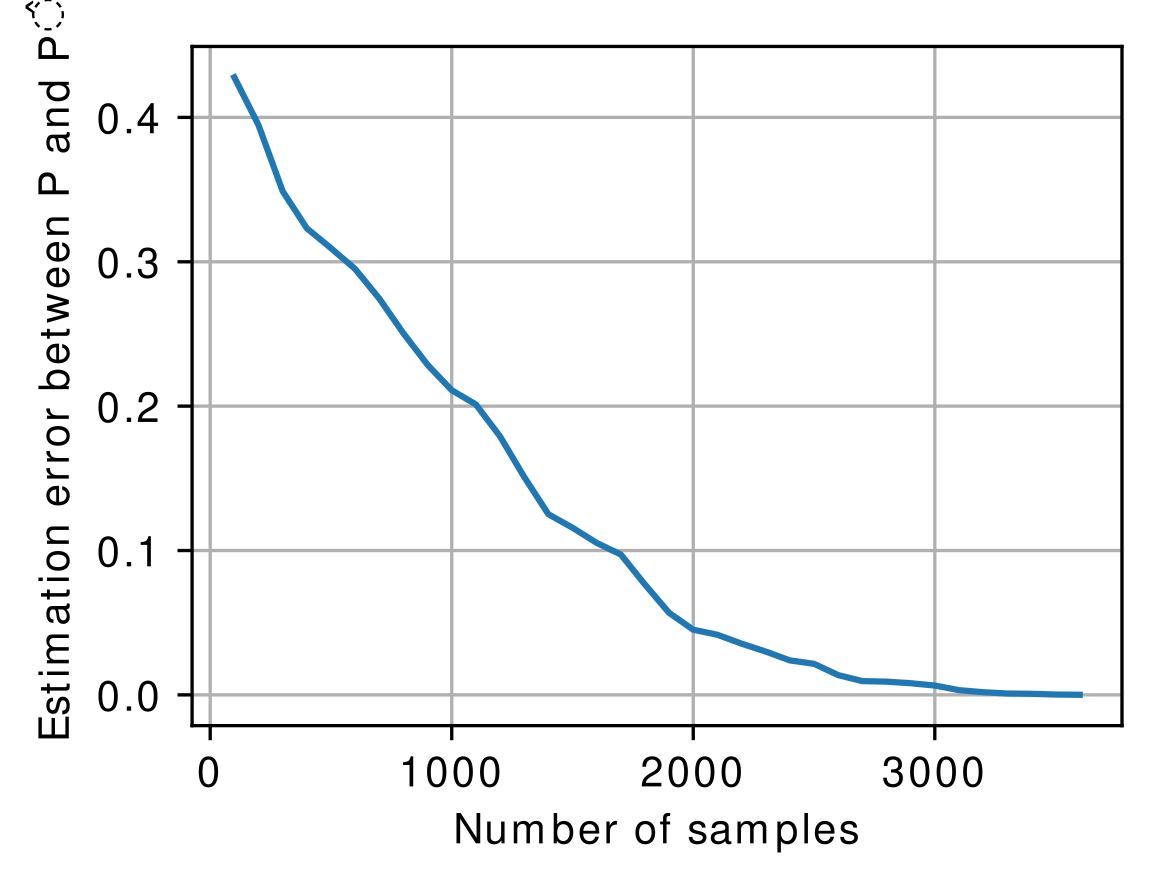}
    \caption{Estimation error between $\mathbf{P}$ and $\mathbf{\hat{P}}$ vs number of samples collected using sample averaging}
    \label{Fig:sample_complexity}
\end{figure} 

We demonstrate the quality of the estimation of the PTT $\mathbf{\hat{P}}$ as a function of the total number of samples collected from the original Markovian environment $\mathcal{M}^{(1)}$ in Algorithm \ref{Algorithm:sampling_creating_multiple_ptts} in Fig.\ref{Fig:sample_complexity} using sample averaging \cite{sample_averaging_2}. We employ the model-4 with a network size of 5000. We measure the quality of the estimation via the $l_2$ norm as follows:
\begin{align}
    \text{estimation error} = \frac{1}{|\mathcal{A}|}\sum_{a \in \mathcal{A}} \|\mathbf{P}_a - \mathbf{\hat{P}}_a\|_2.
\end{align}

Numerical simulations show that an estimation error below 0.05, which requires around 1800 samples, can produce a sufficiently accurate $\hat{\mathbf{P}}$ in Algorithm \ref{Algorithm:sampling_creating_multiple_ptts}, leading to less than 10\% APE in Algorithm \ref{Algorithm: ensemble_link_learning}. We observe that a relatively small number of samples is sufficient for estimating $\mathbf{P}$ to achieve a small APE, which can be attributed to the sparsity and several structural properties of the underlying PTT of the network model (as in \cite{talha_eusipco}). These features also allow us to use shorter trajectories ($l \approx 10$) and select random initial states more frequently in Algorithm \ref{Algorithm:sampling_creating_multiple_ptts}, thereby further accelerating the exploration of the entire state-space.

\begin{figure}[t]
    \centering
    \subfloat[Weights across iterations \label{Fig:convergence_weights}]
    {{\includegraphics[width=4.47cm]{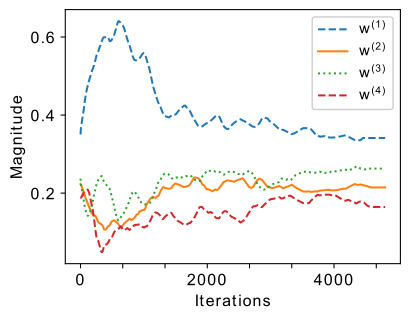}}}
    \subfloat[Simulation error variance vs upper bounds on error variance\label{fig:bound_simulation} ]{{\includegraphics[width=4.4cm]{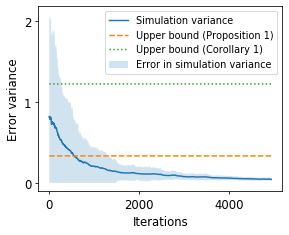}}}
    
    \subfloat[Simulation error expectation \label{Fig:error_expectation}]
    {{\includegraphics[width=4.3cm]{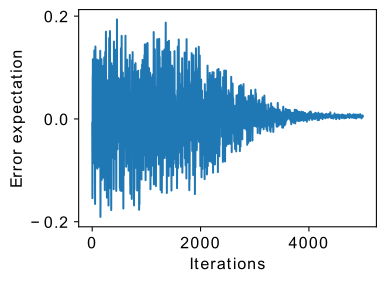}}} 
    \subfloat[Simulation error variance vs $K$ \label{fig:bound_func_K} ]{{\includegraphics[width=4.47cm]{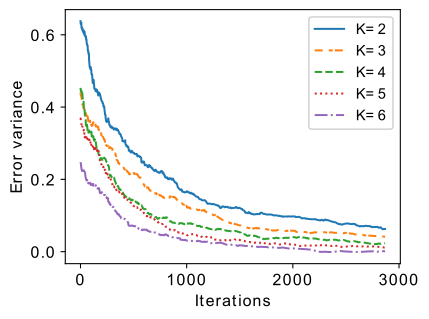}}}
    
    \subfloat[$Q$-function norm difference vs $n$ \label{fig:q_func_diff_bound}]
    {{\includegraphics[width=4.3cm]{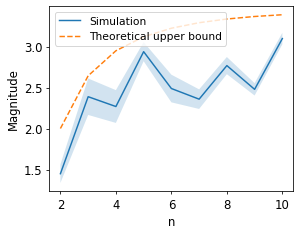}}}
    \caption{Numerical validation of assumptions for theoretical results.}
\end{figure}

\subsection{Convergence of weights}\label{subsec: weights_convergence}
We herein demonstrate the changing weights over time using the simulation settings from Section \ref{Subsec:APE} in Fig.\ref{Fig:convergence_weights}. There is a sharp increase in $\mathbf{w}^{(1)}$ at the beginning because $\mathcal{M}^{(1)}$ is the original environment, and there are not enough samples to capture the higher-order relationships. Moreover, it is not clear which $\mathcal{M}^{(n)}$ provides the most useful samples as the weights $\mathbf{w}^{(n)}$ for $n > 1$ keep changing. As iterations continue, $\mathbf{w}^{(1)}$ decreases up to some point, but $\mathcal{M}^{(1)}$ remains the most useful environment. The weights $\mathbf{w}^{(n)}$ for $n > 1$ increase and converge to a fixed value. The final magnitudes of $\mathbf{w}^{(n)}$ are non-monotonic across $n$ ($\mathbf{w}^{(1)} > \mathbf{w}^{(3)} > \mathbf{w}^{(2)} > \mathbf{w}^{(4)}$), validating the partial ordering of Proposition \ref{proposition-4}. We note that similar weight patterns can be also shown across different network parameters and models, although (i) the final values of the weights, (ii) the order of environment utilities, and (iii) the iteration index at which the weights converge may vary.

\subsection{Numerical validation of propositions}

In this section, we simulate the results in the propositions and compare theoretical results with simulation results. The same simulation settings in Section \ref{Subsec:APE} are employed.

We compute the simulation variance of the $Q$-function errors numerically as follows:
\begin{align}
    {\mathbb{V}[\mathcal{E}_t(s,a)] \approx \frac{1}{2\Delta_t}\sum_{t' = t - \Delta_t}^{t + \Delta_t} \mathcal{E}_t(s,a)^2 \text{--} \Big[\frac{1}{2\Delta_t}\sum_{t' = t - \Delta_t}^{t + \Delta_t} \mathcal{E}_t(s,a)\Big]^2},
\end{align}
with $\Delta_t \ll t$. The upper bounds on the variance from Proposition \ref{proposition-1} and Corollary \ref{corollary0} (with $u_t$ = 0.5 and $\lambda$ = 1) and the simulation variance {(with $(s,a)=(6,2)$ and $\Delta_t = 20$)} are shown in Fig.\ref{fig:bound_simulation}, where the blue-shaded region represents the simulation error. As iterations continue, the simulation variance becomes smaller than the upper bound from Proposition \ref{proposition-1}; hence, using a time-varying $u_t$ produces more accurate results. The simulation variance eventually converges to zero, which is in line with Corollary \ref{corollary}. Herein, $\lambda$ is chosen to be 1, and changing $\lambda$ may affect the initial error variance and its decay rate. Furthermore, the simulation bias (i.e., the expectation of $\mathcal{E}_t(6,2)$) is shown in Fig.\ref{Fig:error_expectation}, where the blue-shaded area represents the simulation error. The simulation bias eventually converges to zero, which is consistent with Fig.\ref{fig:bound_simulation} and shows the accuracy of the assumption (\ref{Equ: simplified_distribution_assumption}).

The simulation error variances across different $K$ are shown in Fig.\ref{fig:bound_func_K}. For each $K$, the simulation variance reduces and converges to zero eventually. The convergence may require more than 3000 iterations, particularly with small $K$, due to slower convergence as indicated by the theoretical runtime complexity in Section \ref{Subsec:runtime_complexity}. Hence, small non-zero error variances are observed at $t=3000$. At a fixed iteration index, the simulation variance also reduces as $K$ increases, which is in line with Proposition \ref{proposition-2}. We also observe the diminishing return of increasing $K$. In particular, the reduction in error variance from $K$ systems to $K+1$ systems becomes less significant as $K$ increases. This implies that increasing $K$ may not always yield substantial improvements because increasing $K$ results in new environments that exhibit high structural similarity with those of lower; the corresponding PTTs converge to fixed tensors. Consequently, sampling from these new environments may not provide novel samples. Furthermore, a larger $K$ leads to a reduction in the weights assigned to existing environments, which diminishes the impact of each individual environment on the ensemble estimate.

Both the upper bound on the $l_2$ norm difference between the $Q$-function vectors of the original environment and $n^{th}$ environment under the output policy of Algorithm \ref{Algorithm: ensemble_link_learning} as a function of $n \in [2,10]$ (from Proposition \ref{proposition-3}) {and the actual norm difference} is illustrated in Fig.\ref{fig:q_func_diff_bound}, where the blue curve is the mean, and the blue shaded area is the standard deviation of the results over 20 simulations. {The norm difference, which represents the utility of the $n^{th}$ order environment, exhibits a non-monotonic behavior across different $n$, and the results align with Proposition \ref{proposition-4}. We also observe that the upper bound converges as $n$ increases, yet the tightness of the bound depends on $n$.}

\begin{figure}[t]
    \centering
    \subfloat[The distribution of $\mathbf{Q}^{(1)}_t(6,2)$ and normal fit \label{Fig:Q_func_dist}]
    {\includegraphics[width=4.3cm]{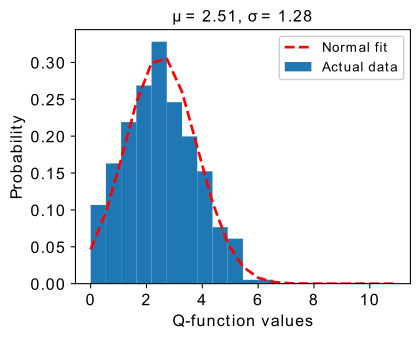}}\hspace{2pt}
    \subfloat[The distribution of $\mathcal{X}^{(2)}_t(6,2)$ and normal fit\label{fig:Q_func_error_dist} ]{{\includegraphics[width=4.25cm]{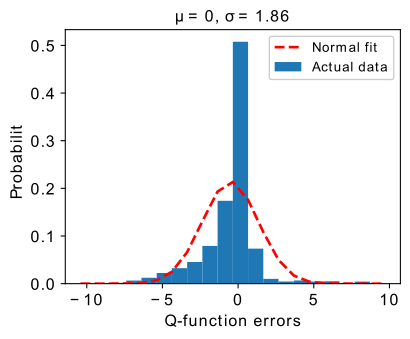}}}

    \subfloat[The distribution of $\mathbf{Q}^{it}_t(6,2)$ and normal fit \label{Fig:Q_func_it_dist} ]{\includegraphics[width=4.3cm]{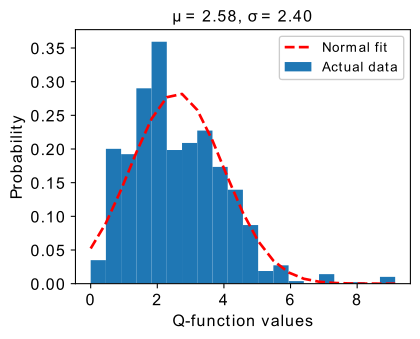}}\hspace{2pt}
    \subfloat[Partial ordering of $Q$-functions for $s=6$ \label{Fig:partial_ordering}]
    {{\includegraphics[width=4.2cm]{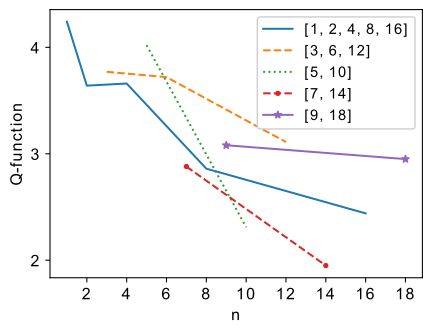}}}

    \subfloat[Normal fit to $\mathcal{X}^{(n)}_t(6,2)$ for different $n$ \label{Fig:normal_fit}]
    {{\includegraphics[width=4.2cm]
    {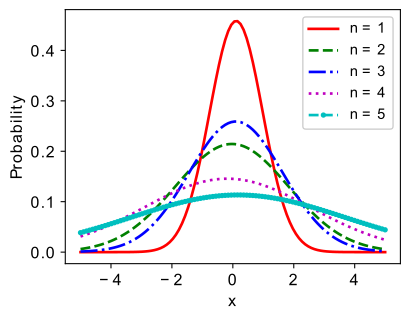}}}\hspace{2pt}
    \subfloat[The averaged distance correlation across different $n$ \label{Fig:corr_coef}]
    {{\includegraphics[width=4.3cm]{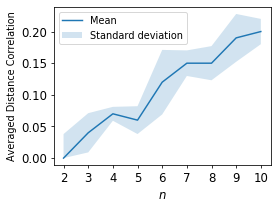}}} 
    \caption{Numerical validation of assumptions in theoretical results.}
\end{figure}

The estimated distributions of $\mathbf{Q}^{(1)}_t(s,a)$, $\mathcal{X}^{(n)}_t(s,a)$ and $\mathbf{Q}^{it}_t(s,a)$ over time {(until convergence)} for $(s,a,n)=(6,2,2)$ are shown in Fig.\ref{Fig:Q_func_dist}, Fig.\ref{fig:Q_func_error_dist} and Fig.\ref{Fig:Q_func_it_dist}, respectively. These distributions can be accurately modeled using normal distributions with corresponding means and variances. Furthermore, we observe that $\mathcal{X}^{(2)}_t(6,2)$ has zero-mean, which shows the accuracy of the zero-mean assumption (\ref{Equ: distribution_assumption}) and thus also the relaxed assumption (\ref{Equ: simplified_distribution_assumption}). Similarly, Fig.\ref{Fig:normal_fit}, which shows the normal fitting of Fig.\ref{fig:Q_func_error_dist} duplicated for various values of $n$, demonstrates that the $Q$-function errors of different environments can be well-modeled by zero-mean normal distributions with different variances. This figure further validates the assumptions (\ref{Equ: distribution_assumption}) and (\ref{Equ: simplified_distribution_assumption}). Similar results are also achieved for different $(s,a)$ pairs.

The $Q$-function of $s=6$ of the $n^{th}$ environment under the output policy of Algorithm \ref{Algorithm: ensemble_link_learning} is shown in Fig.\ref{Fig:partial_ordering} for $n \in [1,20]$. The simulation is carried out with $\gamma = 1 - 10^{-5}$. We observe that partial orderings given in Proposition \ref{proposition-4} hold. For example we have $\mathbf{Q}^{(1)}_{\hat{\pi}}(6) \geq \mathbf{Q}^{(2)}_{\hat{\pi}}(6) \geq \mathbf{Q}^{(4)}_{\hat{\pi}}(6) \geq \mathbf{Q}^{(8)}_{\hat{\pi}}(6)$ and $\mathbf{Q}^{(3)}_{\hat{\pi}}(6) \geq \mathbf{Q}^{(6)}_{\hat{\pi}}(6) \geq \mathbf{Q}^{(12)}_{\hat{\pi}}(6)$ but the relationships between the $Q$-functions of different partial ordering groups are non-monotonic across $n$ such as $\mathbf{Q}^{(6)}_{\hat{\pi}}(6) \geq \mathbf{Q}^{(4)}_{\hat{\pi}}(6)$ and $\mathbf{Q}^{(7)}_{\hat{\pi}}(6) \geq \mathbf{Q}^{(16)}_{\hat{\pi}}(6)$. Extensive simulations demonstrate that similar orderings hold for all states $s$.

To assess the practicality of the assumption employed in Proposition \ref{proposition-1}, we compute the averaged distance correlation (ADC) \cite{distance_correlation_ref} between $\mathcal{X}^{(n)}_{t_1}(6,2)$ and $\mathcal{X}^{(n)}_{t_2}(6,2)$ with $n \in [2,10]$ and averaging the results over all $t_1 \neq t_2 \in [0,5000]$. ADC is particularly used because it captures both linear and non-linear correlations, is robust to outliers, and does not assume any particular distribution of the variables like Pearson's correlation coefficient. The results are shown in Fig.\ref{Fig:corr_coef}. For example, for $n=2$, the ADC is almost 0, indicating that the $Q$-function errors of the $2^{nd}$ environment at different times are independent. When $n$ is modestly large, the ADC is sufficiently small to infer that the correlation is not statistically significant, and the independence assumption holds. However, for very large values of $n$, weak correlations may emerge, suggesting that very large values of $n$ might not provide the best performance.

There are several reasons that explain the independence of the $Q$-function errors at different times: (i) Each environment is Markovian; hence, the errors in $Q$-function estimates can be considered independent. (ii) Each environment is stationary (with constant transition probabilities and reward distributions over time), and the errors in $Q$-function estimates can be assumed independent across different time steps. (iii) The epsilon-greedy policy introduces randomness during the action selection process. (iv) The time-varying learning rate ($\alpha_t$), epsilon probability ($\epsilon_t$) and the update ratio ($u_t$) introduces time-dependent randomness and variability.

\subsection{Parameter Tuning}\label{subsec: Hyper-parameter tuning}
The parameters of Algorithm \ref{Algorithm: ensemble_link_learning} require fine-tuning across various settings. We discuss how to select $v, l, K, \alpha_t, \epsilon_t$ for small ($|\mathcal{S}| \leq 10^3$), modest-sized ($|\mathcal{S}| \in [10^3,10^4]$), and large networks ($|\mathcal{S}| \geq 10^4$). Table \ref{Table:hyper_parameters} summarizes the parameters that yield near-optimal APE values in our numerical results. However, optimal parameter selection requires cross-validation.

The number of visits ($v$) needed for each state-action pair is independent of network size. The proposed algorithm allows a small $v$, around $v \approx 40$, to achieve near-optimal performance. Additionally, the proposed algorithm enables us to have short trajectories $(l)$ while ensuring: (i) near-optimal APE performance, (ii) minimized runtime and computations per trajectory, (iii) sufficient capture of samples from neighboring states, (iv) preservation of initial state importance despite discounting in long trajectories, and (v) prevention of following redundant paths and loops that provide no new samples.

Increasing the number of Markovian environments ($K$) reduces the average runtime complexity of the algorithm. However, there are drawbacks to consider: (i) emphasizing high-order node relationships may lead to the loss of low-order node relationships, (ii) $\mathbf{\hat{P}}^{(n)}$ converges to a fixed tensor as $n$ increases, rendering samples from corresponding {SMEs} redundant and potentially degrading performance, {(iii) a diminishing return is observed in the error variance (as shown in Fig.\ref{fig:bound_func_K}),} and (iv) memory requirements increase. Hence, $K$ should be small enough to avoid these drawbacks, while also increasing with the network size to prevent a rise in other parameters (specifically $v, l$) and reduce runtime complexity.

The learning rate ($\alpha_t$) must adhere to the convergence conditions of $Q$-learning \cite{q-learning-convergence} and have a suitable decay to adjust the learning speed. We assume the form $\alpha_t = \frac{1}{1+\sfrac{t}{c_1}}$, where $c_1 > 0$ determines the decay rate and should increase with $|\mathcal{S}|$, $v$, $l$, and $K$. On the other hand, the parameter $\epsilon_t$ is essential for balancing exploration and exploitation. We use the form $\epsilon_t = \max((c_2)^t, c_3)$, where $c_2 > 0$ adjusts the decay rate, and $0 < c_3 \ll 1$ determines the minimum exploration probability. As the system parameters increase, $c_2$ should also increase to ensure sufficient exploration. Furthermore, $c_3$ should be small and positive, allowing for exploration with a low probability when the policy is nearly converged.

\begin{table}[t]
\centering
\setlength{\tabcolsep}{2pt}
\begin{tabular}{|c| c| c| c| c|} 
 \hline
 Params & Small networks & Modest-sized networks & Large networks\parbox{0pt}{\rule{0pt}{0.5ex+\baselineskip+1ex}}\\
 \hline\hline
  $l$ & $l \in [1,5]$ & $l \in [5,10]$ & $l \in [10,20]$
  \parbox{0pt}{\rule{0pt}{0.5ex+\baselineskip+1ex}}\\
 \hline
  $K$ & $K \in \{2,3\}$ & $K \in \{3,4,5\}$ & $K \in \{5,6,7,8\}$  \parbox{0pt}{\rule{0pt}{0.5ex+\baselineskip+1ex}}\\
 \hline
 $\alpha_t$ & $c_1 \in \{10^2, 5\cdot10^2\}$ & $c_1 \in \{10^2, 10^3\}$ & $c_1 \in \{10^3, 10^4\}$
 \parbox{0pt}{\rule{0pt}{1.5ex+\baselineskip+0ex}}\\
 \hline
 \parbox{0pt}{\rule{0pt}{1.5ex+\baselineskip+2ex}}
 $\epsilon_t$ & \begin{tabular}{l} $c_2 \in \{0.9,0.95\}$ \\ $c_3 \in \{0.01,0.1\}$ \end{tabular} & \begin{tabular}{l} $c_2 \in \{0.95,0.99\}$ \\ $c_3 \in \{0.01,0.05\}$ \end{tabular} & \begin{tabular}{l} $c_2 \in \{0.99, 0.999\}$ \\ $c_3 \in \{0.005,0.01\}$ \end{tabular}  \parbox{0pt}{\rule{0pt}{0.5ex+\baselineskip+2ex}}\\
 \hline
 $u_t$ & $c_4 \in \{10^2,5\cdot10^2\}$ & $c_4 \in \{10^2,10^3\}$ & $c_4 \in \{5 \cdot10^3,10^4\}$  \parbox{0pt}{\rule{0pt}{0.5ex+\baselineskip+1ex}}\\
 \hline
\end{tabular}
\caption{Optimized hyper-parameter values of Algorithm \ref{Algorithm: ensemble_link_learning}.}
\label{Table:hyper_parameters}
\end{table}

The parameter $u_t$ is time-varying such that $u_t \xrightarrow{t \rightarrow \infty} 1$ to balance between exploration and exploitation as explained after Corollary \ref{corollary}. To this end, there are several possible choices for the structure of $u_t$, including $u_t = 1 - e^{\frac{-t}{c_4}}$ and $u_t = 1 - \frac{1}{1 + \frac{t}{c_4}}$ with $c_4 > 0$ or $u_t = 1 - (c_4)^t$ with $c_4 \in (0,1)$. Numerical results show that the performance of the algorithm is maximized for the network models we considered with $u_t = 1 - e^{\frac{-t}{c_4}}$ with optimized $c_4$. In particular, this specific choice leads to 15\% less APE with 20\% less runtime complexity than the other two alternative choices using the simulation settings in Section IV-B. Additionally, this choice of $u_t$ maintains stable performance even when network parameters change significantly, making it a robust choice.

\subsection{Memory complexity}\label{subsec: Memory_complexity}
The memory needs of the proposed algorithm increase with $K$ as the Q-functions of $K$ different Markovian environments need to be stored in tables. To alleviate this challenge, several strategies can be employed, including a linear and non-linear approximation of $Q$-functions or state aggregation methods \cite{q_learning_linear_approx, deep_q_networks, q_learning_aggregation, q_learning_state_abst}. In our prior work \cite{talha_eusipco}, we proposed a special state-action aggregation algorithm to handle the increasing memory complexity of a model-free learning algorithm, similar to Algorithm \ref{Algorithm: ensemble_link_learning}, where $Q$-functions of state-action pairs that incur the same cost are grouped into a single $Q$-function. The algorithm is applicable to networks that have bounded and smoothly changing cost functions (as in model-2, 3, and 4) and can significantly reduce the memory needs while negligibly increasing the APE. 

We note that more sophisticated function approximation algorithms can also be designed to work with any kind of network; however, they require the design of optimal basis functions, selection of appropriate features, or design of aggregation schemes, which are out of the scope of this paper.

\section{Conclusions}\label{sec:conclusion}

In this paper, we presented a novel ensemble $Q$-learning algorithm to overcome the performance and complexity challenges of the original $Q$-learning across large MDPs. The proposed algorithm employs multiple $Q$-function estimators on multiple, distinct, synthetically created, and structurally related Markovian environments that run at different time-scales and fuses the outputs into a single estimate based on an adaptive weighting mechanism using Jensen-Shannon divergence. Extensive simulations across a variety of real-world networks show that the proposed algorithm produces a near-optimal policy with significantly lower complexity and outperforms other $Q$-learning algorithms in terms of accuracy and complexity. Several theoretical upper bounds on the error variance are given, and it is shown that the error variance decreases with the number of Markovian environments. An upper bound on the difference between the $Q$-functions of different environments is also given. In the end, it is shown that the simulation results closely follow the theoretical results. Our analysis provides initial insights and justifications for the advantages of leveraging multiple synthetic Markovian environments in ensemble reinforcement learning. However, there are areas for further improvement and understanding of the nEQL algorithm. Specifically, our ongoing work focuses on devising computationally efficient sampling methods to ensure good data coverage (as described in \cite{coverage_paper}) across all Markovian environments. Additionally, we are exploring alternative approaches to construct synthetic systems that go beyond relying on $n$-hop environments to further improve the generalizability and practicality of our algorithm. Finally, we are working on the extension of the proposed algorithm to continuous state-spaces.

\section{Acknowledgements}\label{sec:acknowledgements}
The authors thank Prof. Tara Javidi of UCSD for ongoing discussions regarding this work and for suggesting that we consider multi-hop transition matrices based on our work on co-link approximations \cite{colink_journal}.

\appendix
\subsection{Proof of Proposition 1}\label{Appendix: proposition_1}

The following expressions are valid for all $(s,a)$ pairs; hence, we drop the $(s,a)$ notation for simplicity. We first prove the expectation.
\begin{align}
    \lim_{t\rightarrow \infty}\mathcal{E}_t &= \lim_{t\rightarrow \infty}\mathbf{Q}_{t}^{it} - \mathbf{Q}^{*}.\label{Equ:def_0}\\
    &= \lim_{t\rightarrow \infty}(1-u)\sum_{i=0}^{t-1}u^{t-i-1}\sum_{n=1}^K\mathbf{w}_i^{(n)}\mathbf{Q}_i^{(n)} - \mathbf{Q}^{*}.\label{Equ:def_1}\\
    &= \lim_{t\rightarrow \infty}(1-u)\sum_{i=0}^{t-1}u^{t-i-1}\sum_{n=1}^K\mathbf{w}_i^{(n)}(\mathbf{Q}_i^{(n)} - \mathbf{Q}^{*}).\label{Equ:def_2}\\
    &= \lim_{t\rightarrow \infty}(1-u)\sum_{i=0}^{t-1}u^{t-i-1}\sum_{n=1}^K\mathbf{w}_i^{(n)}\mathcal{X}^{(n)}_i, \label{Equ:def_2222}
\end{align}
where (\ref{Equ:def_1}) follows from the explicit expression for $\mathbf{Q}_{t}^{it}$, which can be obtained by repeatedly plugging the expression of $\mathbf{Q}_{t-1}^{it}$ in $\mathbf{Q}_{t}^{it}$ in line 11 in Algorithm \ref{Algorithm: ensemble_link_learning}, (\ref{Equ:def_2}) follows from the fact that $\sum_{n=1}^K\mathbf{w}_t^{(n)} = 1$ for all $t$, and $(1-u)\sum_{i=0}^{t-1}u^{t-i-1} = 1$ as $t \rightarrow \infty$, and (\ref{Equ:def_2222}) follows from (\ref{Equ: distribution_assumption}). If we take the expectation of both sides:
\begin{align}
    \lim_{t\rightarrow \infty}\mathbb{E}[\mathcal{E}_t] &= \lim_{t\rightarrow \infty}(1-u)\sum_{i=0}^{t-1}u^{t-i-1}\sum_{n=1}^K\mathbf{w}_i^{(n)}\mathbb{E}[\mathcal{X}^{(n)}_i] = 0\label{Equ:exp_1}
\end{align}
which follows from the linearity of expectation and (\ref{Equ: distribution_assumption}).

We now prove the upper bound on the variance.
$\lim_{t\rightarrow \infty}\mathbb{V}[\mathcal{E}_t] = $
\begin{align}
     &= \lim_{t\rightarrow \infty}\mathbb{V}\Big[(1-u)\sum_{i=0}^{t-1}u^{t-i-1}\sum_{n=1}^K\mathbf{w}_i^{(n)}\mathcal{X}^{(n)}_i\Big].\label{Equ:var_1}\\
    &= \lim_{t\rightarrow \infty}(1-u)^2\Big[\sum_{i=0}^{t-1}u^{2(t-i-1)}\big[ \sum_{n=1}^K{(\mathbf{w}_i^{(n)})}^2\mathbb{V}[\mathcal{X}^{(n)}_i] \nonumber \\ +& 2\sum_{n=1}^K\sum_{m=n+1}^K\mathbf{w}_i^{(n)}\mathbf{w}_i^{(m)}\operatorname{Cov}(\mathcal{X}^{(n)}_i, \mathcal{X}^{(m)}_i)\big]\Big]. \label{Equ:var_2}\\
    &\leq \lim_{t\rightarrow \infty}(1-u)^2\Big[\sum_{i=0}^{t-1}u^{2(t-i-1)}\big[ \sum_{n=1}^K{(\mathbf{w}_i^{(n)})}^2\mathbb{V}[\mathcal{X}^{(n)}_i] \nonumber \\ +& 2\sum_{n=1}^K\sum_{m= n+1}^K\mathbf{w}_i^{(n)}\mathbf{w}_i^{(m)}\sqrt{\mathbb{V}[\mathcal{X}^{(n)}_i]\mathbb{V}[\mathcal{X}^{(m)}_i}]\big]\Big].\label{Equ:var_4}\\
    &\leq \lim_{t\rightarrow \infty}(1-u)^2\Big[\sum_{i=0}^{t-1}u^{2(t-i-1)}\big[ \sum_{n=1}^K\mathbf{w}_i^{(n)}\mathbb{V}[\mathcal{X}^{(n)}_i] \nonumber \\ +& 2\sum_{n=1}^K\sum_{m=1}^K\mathbf{w}_i^{(n)}\mathbf{w}_i^{(m)}\sqrt{\mathbb{V}[\mathcal{X}^{(n)}_i]\mathbb{V}[\mathcal{X}^{(m)}_i}]\big]\Big].\label{Equ:var_5}\\
    &\leq \lim_{t\rightarrow \infty}(1-u)^2\Big[\sum_{i=0}^{t-1}u^{2(t-i-1)}\big[ \sum_{n=1}^K\mathbf{w}_i^{(n)}\frac{\lambda^2}{3} \nonumber \\ +& 2\sum_{n=1}^K\sum_{m=1}^K\mathbf{w}_i^{(n)}\mathbf{w}_i^{(m)}\frac{\lambda^2}{3}\big]\Big].\label{Equ:var_6}\\
    &\leq \lim_{t\rightarrow \infty}(1-u)^2\Big[\sum_{i=0}^{t-1}u^{2(t-i-1)}\lambda^2\Big]. \label{Equ:var_7}\\
    &\leq \frac{(1-u)}{(1+u)}\lambda^2,\label{Equ:var_8}
\end{align}
where (\ref{Equ:var_1}) follows from (\ref{Equ:def_2222}), (\ref{Equ:var_2}) follows from the properties of the variance operator and the independence assumption, (\ref{Equ:var_4}) follows from the Cauchy-Schwarz inequality for the variance, (\ref{Equ:var_5}) follows from the fact that $\mathbf{w}_t^{(n)} \leq 1$ and dropping the constraint in the second summation, (\ref{Equ:var_6}) follows from (\ref{Equ: distribution_assumption}) and $\lambda = \max\limits_n\lambda_n$, (\ref{Equ:var_7}) follows from the fact that $\sum_{n=1}^K\mathbf{w}_t^{(n)} = 1$ for all $t$, and (\ref{Equ:var_8}) follows from the infinite geometric sum formula and $u \in (0,1)$.

\subsection{Proof of Corollary 1}\label{Appendix: corollary_1}

We can show the expectation result as follows:
\begin{align}
    \lim_{t\rightarrow \infty}\mathbb{E}[\mathcal{E}_t] &= \lim_{t\rightarrow \infty}(1-u)\sum_{i=0}^{t-1}u^{t-i-1}\sum_{n=1}^K\mathbf{w}_i^{(n)}\mathbb{E}[\mathcal{X}^{(n)}_i].\label{Equ:modified_1}\\
    &= \lim_{t\rightarrow \infty}(1-u)\sum_{i=0}^{t-1}u^{t-i-1}\sum_{n=1}^K\mathbf{w}_i^{(n)}\mu_n. \label{Equ:modified_2}\\
    &= 0,\label{Equ:modified_3}
\end{align}
where (\ref{Equ:modified_1}) follows from (\ref{Equ:exp_1}), (\ref{Equ:modified_2}) follows from (\ref{Equ: simplified_distribution_assumption}), and (\ref{Equ:modified_3}) follows from the condition in Corollary 1. The proof for the result on the upper bound of the variance in Proposition \ref{proposition-1} remains unchanged.

\subsection{Proof of Proposition 2}\label{Appendix: proposition_2}
We firstly bound the weight $\mathbf{w}^{(n)}$ into a tighter interval than [0,1]. We first do the calculations for $n \neq 1$. The maximum value of $\mathbf{w}^{(n)}$ is obtained when $\mathbf{\hat{Q}}^{(1)} = \mathbf{\hat{Q}}^{(n)}$ and $\mathbf{\hat{Q}}^{(i)}$ is maximally different than $\mathbf{\hat{Q}}^{(1)}$ for all $i \neq n$. In this case, $\mathbf{w}^{(n)} = 1$, and $\mathbf{w}^{(i)} = 0$ for all $i \not\in \{n,1\}$. When we apply the softmax operator on the $\mathbf{w}$, we obtain the following:
\begin{align}
    \mathbf{w}^{(n)} &= \frac{e}{2e + (K-2)} \leq \frac{e}{K}.\label{Equ: max_value_weight_1}
\end{align}
which follows as there are $K > 1$ different weights, and $\mathbf{w}^{(1)}=1$.
On the other hand, the maximum value of $\mathbf{w}^{(1)}$ is obtained when $\mathbf{\hat{Q}}^{(i)}$ is maximally different than $\mathbf{\hat{Q}}^{(1)}$ for all $i \neq 1$. In this case, $\mathbf{w}^{(1)} = 1$, and $\mathbf{w}^{(i)} = 0$ for all $i \neq 1$. When we apply the softmax on $\mathbf{w}$, we obtain the following:
\begin{align}
    \mathbf{w}^{(1)} &= \frac{e}{e + (K-1)} \leq \frac{e}{K}.\label{Equ: max_value_weight_3}
\end{align}

Combining (\ref{Equ: max_value_weight_1}) and (\ref{Equ: max_value_weight_3}), the following holds for all $n$:
\begin{align}
    \mathbf{w}^{(n)} \leq \frac{e}{K}.\label{Equ: max_value_weight_main}
\end{align}

We note that this bound is useful for $K > 2$. On the other hand, the minimum value of $\mathbf{w}^{(n)}$ for $n \neq 1$ is obtained when $\mathbf{\hat{Q}}^{(1)}$ is maximally different than $\mathbf{\hat{Q}}^{(n)}$ and $\mathbf{\hat{Q}}^{(i)} = \mathbf{\hat{Q}}^{(1)}$ for all $i \neq n$. In this case, $\mathbf{w}^{(n)} = 0$, and $\mathbf{w}^{(i)} = 1$ for all $i \neq n$. When we apply the softmax operator on the $\mathbf{w}$, we obtain the following:
\begin{align}
    \mathbf{w}^{(n)} &= \frac{1}{1 + e(K-1)} \geq \frac{e^{-1}}{K}.\label{Equ: min_value_weight_1}\end{align}
which follows because $K > 1$. Similarly, the minimum value of $\mathbf{w}^{(1)}$ is obtained when $\mathbf{\hat{Q}}^{(i)} = \mathbf{\hat{Q}}^{(1)}$ for all $i \neq 1$. In this case, $\mathbf{w}^{(i)} = 1$ for all $i$. When we apply the softmax operator on the $\mathbf{w}$, we obtain the following:
\begin{align}
    \mathbf{w}^{(1)} &= \frac{1}{K}.\label{Equ: min_value_weight_3}
\end{align}

Combining (\ref{Equ: min_value_weight_1}) and (\ref{Equ: min_value_weight_3}), the following holds for all $n$:
\begin{align}
    \mathbf{w}^{(n)} \geq \frac{1}{K}.\label{Equ: min_value_weight_main}
\end{align}

Combining (\ref{Equ: max_value_weight_main}) and (\ref{Equ: min_value_weight_main}), the weights $\mathbf{w}^{(n)}$ can be shown to take values in the following interval: $\mathbf{w}^{(n)} \in [\frac{1}{K},\frac{e}{K}].$ Then, the term $\sum_{n=1}^K{(\mathbf{w}_i^{(n)})}^2$ in (\ref{Equ:var_2}) can be upper bounded as:
\begin{align}
    \sum_{n=1}^K (\mathbf{w}^{(n)}_i)^2 &\leq \sum_{n=1}^K (\frac{e}{K})^2 \leq \frac{e^2}{K}.\label{Equ: approximation_14}
\end{align}

The following expressions are valid for all $(s,a)$ pairs; hence, we drop the $(s,a)$ notation for simplicity. Using (\ref{Equ: approximation_14}), (\ref{Equ: distribution_assumption}), and $\lambda$ = $\max\limits_n\lambda_n$, we can show the following:
\begin{align}
    \mathbb{V}\big[\sum_{n=1}^K\mathbf{w}_i^{(n)}\mathcal{X}^{(n)}_i\big] &\leq \sum_{n=1}^K{(\mathbf{w}_i^{(n)})}^2\mathbb{V}\big[\mathcal{X}^{(n)}_i\big] \leq \frac{e^2}{K}\frac{\lambda^2}{3}. \label{Equ: var_upper_bound}
\end{align}

{Let $u_0 = 0$ and $\bar{u}_i = (1-u_i)\prod_{j=i+1}^{t-1}u_j$. Using the explicit expression for $\mathbf{Q}_{t}^{it}$, which can be obtained by repeatedly plugging the expression of $\mathbf{Q}_{t-1}^{it}$ in $\mathbf{Q}_{t}^{it}$ in line 11 in Algorithm \ref{Algorithm: ensemble_link_learning} and the facts that $\sum_{i=0}^t\bar{u}_i = 1$ and $u_0=0$, the following can be derived:
\begin{align}
    \mathbf{Q}_t^{it} = \sum_{i=0}^{t-1}\bar{u}_i\sum_{n=1}^K\mathbf{w}_i^{(n)}\mathbf{Q}^{(n)}_i, \label{Equ: recursive_Q_it_t_dependent}
\end{align}}

Then we can show the following:
\begin{align}
 \mathbb{V}[\mathcal{E}_t] &=\mathbb{V}\Big[\sum_{i=0}^{t-1}\bar{u}_i\sum_{n=1}^K\mathbf{w}_i^{(n)}\mathcal{X}^{(n)}_i\Big].\label{Equ: var_upper_bound_no_indp_asum_1}\\
 &= \sum_{i=0}^{t-1}\bar{u}_i^2\mathbb{V}[\sum_{n=1}^K\mathbf{w}_i^{(n)}\mathcal{X}^{(n)}_i] + \nonumber\\ &2\sum_{i=0}^{t-1}\sum_{j\neq i}^{t-1}\bar{u}_i\bar{u}_j\operatorname{Cov}(\sum_{n=1}^K\mathbf{w}_i^{(n)}\mathcal{X}^{(n)}_i, \sum_{n=1}^K\mathbf{w}_j^{(n)}\mathcal{X}^{(n)}_j).\label{Equ: var_upper_bound_no_indp_asum_2}\\
 &\leq \sum_{i=0}^{t-1}\bar{u}_i^2\mathbb{V}[\sum_{n=1}^K\mathbf{w}_i^{(n)}\mathcal{X}^{(n)}_i] + \nonumber\\ &2\sum_{i=0}^{t-1}\sum_{j= i+1}^{t-1}\bar{u}_i\bar{u}_j\sqrt{\mathbb{V}(\sum_{n=1}^K\mathbf{w}_i^{(n)}\mathcal{X}^{(n)}_i)\mathbb{V}(\sum_{n=1}^K\mathbf{w}_j^{(n)}\mathcal{X}^{(n)}_j)}.\label{Equ: var_upper_bound_no_indp_asum_3}\\
 &\leq \frac{e^2}{K}\frac{\lambda^2}{3}\Big[\sum_{i=0}^{t-1}\bar{u}_i^2 + 2\sum_{i=0}^{t-1}\sum_{j=i+1}^{t-1}\bar{u}_i\bar{u}_j\Big].\label{Equ: var_upper_bound_no_indp_asum_4}\\
 &\leq \frac{c(\lambda, u)}{K},\label{Equ: var_upper_bound_no_indp_asum_5}
\end{align}
where (\ref{Equ: var_upper_bound_no_indp_asum_1}) follows from (\ref{Equ: recursive_Q_it_t_dependent}), (\ref{Equ: var_upper_bound_no_indp_asum_2}) follows from the properties of the variance operator, (\ref{Equ: var_upper_bound_no_indp_asum_3}) follows from the Cauchy-Schwarz inequality for the variance, (\ref{Equ: var_upper_bound_no_indp_asum_4}) follows from (\ref{Equ: var_upper_bound}), and (\ref{Equ: var_upper_bound_no_indp_asum_5}) follows as $c(\lambda, u)$ is a constant of $K$.

\subsection{Proof of Proposition 3}\label{Appendix: proposition_3}
If the spectral norm of a matrix $\mathbf{A}$ is less than 1 $(\|\mathbf{A}\| < 1)$, then the following can be shown \cite{strang_book}:
\begin{align}
    (\mathbf{I}-\mathbf{A})^{-1} &= \sum_{i=0}^{\infty} \mathbf{A}^i, \label{Equ: matrix_inverse}
\end{align}
where $\mathbf{I}$ is the identity matrix. The $Q$-functions under policy $\hat{\pi}$ can be obtained using Bellman's equation as follows \cite{bertsekas_book}:
\begin{align}
    \mathbf{Q}_{\hat{\pi}} &= (\mathbf{I} - \gamma\mathbf{P}_{\hat{\pi}})^{-1}\mathbf{c}_{\hat{\pi}},\label{Equ: Bellmans_matrix_equ}
\end{align}
where $\mathbf{P}_{\hat{\pi}}$ and $\mathbf{c}_{\hat{\pi}}$ are PTM and cost vectors under policy $\hat{\pi}$, respectively. The matrix inversion in (\ref{Equ: Bellmans_matrix_equ}) can be expressed as an infinite sum using (\ref{Equ: matrix_inverse}) since the spectral norm of $\mathbf{P}_{\hat{\pi}}$ is 1 (as it is a PTM), and when scaled by $\gamma$, its spectral norm becomes strictly less than 1. Then, we can proceed as follows:
\begin{align}
    \mathbf{Q}^{(1)}_{\hat{\pi}}-\mathbf{Q}^{(n)}_{\hat{\pi}} &= (\mathbf{I} - \gamma\mathbf{P}_{\hat{\pi}})^{-1}\mathbf{c}_{\hat{\pi}} - (\mathbf{I} - \gamma\mathbf{P}^{n}_{\hat{\pi}})^{-1}\mathbf{c}_{\hat{\pi}}.\label{Equ: temp_equ_15}\\
    &= \big[\gamma\mathbf{P}_{\hat{\pi}}\sum_{i=0}^\infty (\gamma\mathbf{P}_{\hat{\pi}})^{ni}\sum_{k=0}^{n-2}(\gamma\mathbf{P}_{\hat{\pi}})^k + \nonumber\\& \hspace{70pt} \sum_{j=0}^\infty (\gamma^{nj} - \gamma^j)\mathbf{P}^{nj}_{\hat{\pi}}\big]\mathbf{c}_{\hat{\pi}}. \label{Equ: temp_equ_16}\\
    &\leq \gamma\mathbf{P}_{\hat{\pi}}\sum_{i=0}^\infty (\gamma\mathbf{P}_{\hat{\pi}})^{ni}\sum_{k=0}^{n-2}(\gamma\mathbf{P}_{\hat{\pi}})^k\mathbf{c}_{\hat{\pi}} \label{Equ: temp_equ_17}
\end{align}
where (\ref{Equ: temp_equ_15}) follows from (\ref{Equ: Bellmans_matrix_equ}), (\ref{Equ: temp_equ_16}) follows from the fact that $\gamma\|\mathbf{P}_{\hat{\pi}}\| < 1$, (\ref{Equ: matrix_inverse}) and grouping the common terms in the expansions, and (\ref{Equ: temp_equ_17}) follows from the facts that $\gamma < 1$ and the elements of $\mathbf{P}_{\hat{\pi}}$ and that $\mathbf{c}_{\hat{\pi}}$ are non-negative. If we take the norm of both sides, we obtain:
\begin{align}
    \|\mathbf{Q}^{(1)}_{\hat{\pi}} - \mathbf{Q}^{(n)}_{\hat{\pi}}\| &\leq \|\gamma\mathbf{P}_{\hat{\pi}}\sum_{i=0}^\infty (\gamma\mathbf{P}_{\hat{\pi}})^{ni}\sum_{k=0}^{n-2}(\gamma\mathbf{P}_{\hat{\pi}})^k\mathbf{c}_{\hat{\pi}}\|.\label{Equ: temp_equ_18}\\
    &\leq \gamma\|\mathbf{P}_{\hat{\pi}}\|\sum_{i=0}^\infty (\gamma\|\mathbf{P}_{\hat{\pi}}\|)^{ni}\sum_{k=0}^{n-2}(\gamma\|\mathbf{P}_{\hat{\pi}}\|)^k\|\mathbf{c}_{\hat{\pi}}\|.\label{Equ: temp_equ_19}\\
    &< \gamma\sum_{i=0}^\infty \gamma^{ni}\sum_{k=0}^{n-2}\gamma^k\|\mathbf{c}_{\hat{\pi}}\|.\label{Equ: temp_equ_20}\\
    &< \frac{\gamma}{1-\gamma^n}\frac{1-\gamma^{n-1}}{1-\gamma}\|\mathbf{c}_{\hat{\pi}}\|,\label{Equ: temp_equ_21}
\end{align}
where (\ref{Equ: temp_equ_18}) follows from (\ref{Equ: temp_equ_17}), (\ref{Equ: temp_equ_19}) follows from the upper bound on the norm of a matrix-vector product \cite{strang_book}, (\ref{Equ: temp_equ_20}) follows from the fact that $\gamma\|\mathbf{P}_{\hat{\pi}}\| < 1$, and (\ref{Equ: temp_equ_21}) follows from the finite and infinite geometric sum formulas.

\subsection{Proof of Proposition 4}\label{Appendix: proposition_4}
We analyze $\mathbf{Q}_{\hat{\pi}}^{(n)}$ when $\gamma \rightarrow 1$.
\begin{align}
    \lim_{\gamma\rightarrow1}\mathbf{Q}^{(n)}_{\hat{\pi}} &= \lim_{\gamma\rightarrow1}(\mathbf{I} - \gamma\mathbf{P}^{n}_{\bm{\pi}})^{-1}\mathbf{c}_{\bm{\pi}}\label{Equ: order_chain_1}\\
    &= (\mathbf{I} - \mathbf{P}^{n}_{\bm{\pi}})^{-1}\mathbf{c}_{\bm{\pi}}\label{Equ: order_chain_2}\\
    &= \mathbf{c}_{\bm{\pi}} + \mathbf{P}^{n}_{\bm{\pi}}\mathbf{c}_{\bm{\pi}} + (\mathbf{P}^{n}_{\bm{\pi}})^2\mathbf{c}_{\bm{\pi}} + (\mathbf{P}^{n}_{\bm{\pi}})^3\mathbf{c}_{\bm{\pi}} + ...,
\end{align}
which follows from (\ref{Equ: matrix_inverse}). Then, we can write the following:
\begin{align}
    \lim_{\gamma\rightarrow1}\mathbf{Q}^{(1)}_{\hat{\pi}} &= \mathbf{c}_{\bm{\pi}} + \mathbf{P}_{\bm{\pi}}\mathbf{c}_{\bm{\pi}} + \mathbf{P}^{2}_{\bm{\pi}}\mathbf{c}_{\bm{\pi}} + \mathbf{P}^{3}_{\bm{\pi}}\mathbf{c}_{\bm{\pi}} +  ... \\
    \lim_{\gamma\rightarrow1}\mathbf{Q}^{(2)}_{\hat{\pi}} &= \mathbf{c}_{\bm{\pi}} + \mathbf{P}^{2}_{\bm{\pi}}\mathbf{c}_{\bm{\pi}} + \mathbf{P}^{4}_{\bm{\pi}}\mathbf{c}_{\bm{\pi}} + \mathbf{P}^{6}_{\bm{\pi}}\mathbf{c}_{\bm{\pi}} + ... \\
    \lim_{\gamma\rightarrow1}\mathbf{Q}^{(3)}_{\hat{\pi}} &= \mathbf{c}_{\bm{\pi}} + \mathbf{P}^{3}_{\bm{\pi}}\mathbf{c}_{\bm{\pi}} + \mathbf{P}^{6}_{\bm{\pi}}\mathbf{c}_{\bm{\pi}} + \mathbf{P}^{9}_{\bm{\pi}}\mathbf{c}_{\bm{\pi}} + ...
\end{align}

We observe that $\mathbf{Q}^{(1)}_{\hat{\pi}}$ is the largest since it has all the terms that $\mathbf{Q}^{(n)}_{\hat{\pi}}$ contains for all $n \neq 1$. We also have the following partial orderings in the limiting case ($\gamma\rightarrow1$):
\begin{align}
    \mathbf{Q}^{(2)}_{\hat{\pi}} &\geq \mathbf{Q}^{(4)}_{\hat{\pi}} \geq \mathbf{Q}^{(8)}_{\hat{\pi}} \geq \mathbf{Q}^{(16)}_{\hat{\pi}} ... \\
    \mathbf{Q}^{(3)}_{\hat{\pi}} &\geq \mathbf{Q}^{(6)}_{\hat{\pi}} \geq \mathbf{v}_\pi^{(12)} ... \\
    \mathbf{Q}^{(5)}_{\hat{\pi}} &\geq \mathbf{Q}^{(10)}_{\hat{\pi}} \geq \mathbf{Q}^{(20)}_{\hat{\pi}} ... \\
    \mathbf{Q}^{(7)}_{\hat{\pi}} &\geq \mathbf{Q}^{(14)}_{\hat{\pi}} \geq \mathbf{Q}^{(28)}_{\hat{\pi}} ...,
\end{align}
which follow as the elements of $\mathbf{P}_{\hat{\pi}}$ and $\mathbf{c}_{\hat{\pi}}$ are non-negative.

\bibliographystyle{unsrt}
\bibliography{references.bib}

\end{document}